\documentclass[preprint,12pt,3p]{elsarticle}

\usepackage{amssymb}
\usepackage{amsthm}
 \usepackage{lineno,hyperref}
\usepackage{graphicx}
\usepackage{adjustbox}
\usepackage{numprint}
\usepackage{array}

\newcolumntype{P}[1]{>{\centering\arraybackslash}p{#1}}
\newcolumntype{M}[1]{>{\centering\arraybackslash}m{#1}}

\npdecimalsign{.} 

 \biboptions{sort,square,semicolon}

\begin{document}

\begin{frontmatter}

\title{A joint text mining-rank size investigation of the rhetoric structures of the US Presidents' speeches}

\author[label1,label2]{Valerio Ficcadenti}
\address[label1]{Department of Economics and Law, University of Macerata,
Italy}
\address[label2]{Department of Management, Marche Polytechnic University, Italy}


\ead{ficcadentivalerio@gmail.com}

\author[label1]{Roy Cerqueti\corref{cor1}}
\cortext[cor1]{Corresponding author: Roy Cerqueti, University of
Macerata, Department of Economics and Law. Via Crescimbeni 14 -
I-62100, Macerata, Italy. Tel.: +39 0733 258 3246. Email:
roy.cerqueti@unimc.it} \ead{roy.cerqueti@unimc.it}

\author[label5,label6]{Marcel Ausloos}
\address[label5]{School of Business, University of Leicester, United Kingdom}
\address[label6]{GRAPES,   -- Group of Researchers for Applications of Physics in Economy and Sociology.
Rue de la Belle Jardiniere, 483/0021, B-4031, Liege Angleur, Euroland, Belgium}
\ead{ma683@le.ac.uk}

\begin{abstract}
This work presents a text mining context and its use for a deep
analysis of the messages delivered by the politicians. Specifically,
we deal with an expert systems-based exploration of the rhetoric
dynamics of a large collection of US Presidents' speeches, ranging
from Washington to Trump. In particular, speeches are viewed as
complex expert systems whose structures can be effectively analyzed
through rank-size laws. The methodological contribution of the paper
is twofold. First, we develop a text mining-based procedure for the
construction of the dataset by using a web scraping routine on the
Miller Center website -- the repository collecting the speeches.
Second, we explore the implicit structure of the discourse data by
implementing a rank-size procedure over the individual speeches,
being the words of each speech ranked in terms of their frequencies.
The scientific significance of the proposed combination of
text-mining and rank-size approaches can be found in its flexibility
and generality, which let it be reproducible to a wide set of expert
systems and text mining contexts. The usefulness of the proposed
method and the speech subsequent analysis is demonstrated by the
findings themselves. Indeed, in terms of impact, it is worth noting
that interesting conclusions of social, political and linguistic
nature on how 45 United States Presidents, from April 30, 1789 till
February 28, 2017 delivered political messages can be carried out.
Indeed, the proposed analysis shows some remarkable regularities,
not only inside a given speech, but also among different speeches.
Moreover, under a purely methodological perspective, the presented
contribution suggests possible ways of generating a linguistic
decision-making algorithm.

\end{abstract}


\end{frontmatter}
\section{Introduction}
\label{intro}
The main changes in schools of thought and political arrangements
have been communicated to the public by means of speeches, whence by
different rhetoric structures. Politicians, trying to convince their
own people, and others, about their opinions, have always used the
words as the primary means  \cite{alduy}. The way to build the
rhetoric structures has changed with time. Some aspects of changing in
languages are described in \cite{vejdemo2016semantic}. Evidences of these
changes could be highlighted through text analysis in order to study
classes of words as in \cite{tetlock2008more}.

This paper moves from this premise. It presents a joint text-mining
procedure and rank-size analysis for the description of the
political speeches and their structures. In doing so, we are in line
with a wide strand of literature where texts are viewed as complex
expert systems and text-mining techniques are used for exploring a
number of contexts
\citep{amrit,noh,nassirtoussi2014text,mostafa,oberreuter2013text,berezina,moro,pletsche,westergaard}.

The political environment here analyzed let this paper be
particularly close to \cite{defortuny}, where the authors focus on
the Belgian political uncertainty in 2011 and how the Flemish
newspapers treated it. We here present the paradigmatic case of the
speeches of the US politicians, with special attention to US
Presidents. Indeed, the United States President is one of the most
important people in the world and his speeches are often addressed
to a wide audience. So, there is no doubt about the immanent
relevance of the words pronounced by a President and it is expected
that they bear much influence on the overall economic and social
contexts.

In this work, we deal with analysis of the US presidential speeches
in a larger sense than the mere focus on the meaning of the single
words as it happens for exploring the topics present in a text \cite{griffiths2004finding}. Specifically, we analyze the
constitutive rhetoric structures of the speeches to explore the way
of creating speeches. The basic scientific motivation lies in the
evident accuracy of words selection by the US Presidents and their
collaborators, which leads to the worthiness of understanding if a
speech repeats obsessively few words or is more scattered among a
large numbered of tokens. In this respect, words can be ranked
according to their frequencies, so that each talk represents a
structured system that contains information about the way of
communicating messages to the audience. The inner structure of the
speeches can be inferred by applying a proper best-fit procedure for
a rank-size law.

Our approach is in line with the languages studies investigating
structures of texts, speeches or languages \citep{calude2011we,
ausloos2012generalized,ausloos2012measuring,altmann2012origin,
zeng2012topics}. In our peculiar context, we follow the route
traced by the studies on the connection between political speeches
and Government policies as in \cite{milizia2014out} for the case of
UK.
The contribution \cite{tetlock2007giving} is an example of the
investigation of news' impact on the financial market starting from
premises close to ours. It is also worth mentioning
\cite{lim2002five} where the author identifies the main changes in
the rhetoric of the President Inaugural Addresses and Annual
Messages from George Washington to Bill Clinton. As we will see, the
present paper is radically different from \cite{lim2002five}: (i)
the present dataset is remarkably larger than the
\cite{lim2002five}'s one; (ii) the employed methodology is
different: as we will see in details, we use a rank-size approach,
while the quoted paper adopts the General Inquirer for the specific
assessment of words categories; (iii) the target is different: we
here aim at giving a view of the structure of the speeches through
the tokens frequencies and their ranks as in
\cite{tang2013recognition}, while \cite{lim2002five} pays special
attention to the meaning of the words.

As anticipated, we investigate the rhetorical structure of the US
Presidents by using a different method with respect to the proposals
adopting a computer science perspective. In this framework, we
mention \cite{ATKINSON20134346} for the case of web documents
summarization, \cite{KRAUS201965} where a deep learning algorithm is
used to run a semantic analysis or \cite{saravanan2008automatic},
that deals with a rhetoric study of legal documents. Our work falls
in the context of the processes that have to be implemented before
the application of machine learning models or automated discourse
processing methods (see e.g. \cite{SIDOROV2014853}, where the
authors discuss the data preparation phases). In this specific area,
we highlight the role of the rank-size law parameters as texts
characters keepers, hence reducing the dimension of the textual
data. Under a different perspective, the calibrated parameters of
the rank-size law generate a three-dimensional space which is able
to represent the data (refer to \cite{tang2013recognition} as an
example of efforts done to reach text characters representativeness
in the pre-processing phase of non-supervised clusterization of
corpus). Furthermore, observing the features of the texts from a
rank-size perspective, it is possible to avoid restrictive
assumptions on languages changes and semantic homogeneity, as in
\cite{tang2013recognition}. The followed macro-view approach of the
US Presidents speeches transcripts -- which are observed as if they
constitute a rhetoric framework -- leads to a different
representation of those documents with respect to the classical bag
of word methods.

$N$-grams models are predominant in Natural Language Processing
field. Such methodological proposals are employed to represent texts
in vector spaces where similarity and distance measures are applied
to determine document closeness. In these cases, the analysis of the
words frequencies is the connection between texts processing and
successive analysis. The present research can be effectively
inserted in the set of studies performing pure dataset exploration
with efficient models (see some examples in Table
\ref{textminingtable}) and studies like
\cite{saravanan2006probabilistic,tang2013recognition}, where new
methodological tools are developed). Indeed, we here present an
efficient way to extract the information underlying the speeches
transcripts. In particular, we are far from the filed in which
semantic and syntax are involved and rather deal with a model in the
Information Extraction and Information Theory fields.

During the last years, several stylometric researches have
contributed to the creation of measures to feed the authorship
attribution algorithms (see \cite{stamatatos2009survey} for a wide
review of this theme). In this context, it is important to take into
account the lexical aspects of the writings. Indeed, it is well
known that the assessment of the most frequent words in a writing
foster the identification of its author. Therefore, our procedure
can express some potentialities in authorship attribution since we
deal with the distribution of the terms within documents, by
quantifying the presence of tokens along ranks thanks to the
rank-size parametrization. So, it could be possible to create a
rhetorical profile for each speaker on the basis of the rank-size
analysis of the speeches.

In order to clarify the context in which we are operating we refer
to Table \ref{textminingtable}, where references are related to the
different approaches used in text mining and their applications. In
this way we offer a summarized view of the state of the art.

\begin{table}
\tiny
\begin{tabular}{M{0.15\textwidth} |M{0.5\textwidth} | M{0.25\textwidth}}
\hline
    \textbf{Authors}  &  \textbf{Approach} & \textbf{Description} \\
\hline \cite{altmann2012origin} & Binary transformation of texts;
Correlation analysis& An analysis of long-range correlations in
texts \\ \hline
\cite{amrit} & Supervised classification based on machine learning;
Naive Bayes; Random forest; Support Vector Machine & Identification
of child abuse\\ \hline \cite{ATKINSON20134346} & Lantent Semantic
Analysis; Singular Value Decomposition; Conditional random field
classifier; K-mixture model & New multi-document summarization
methods applied to web-based news \\ \hline
\cite{ausloos2008equilibrium, ausloos2010punctuation, ausloos2012generalized, ausloos2012measuring} & 
Zipf and Zipf-Mandelbrot law; GrassbergerñProccacia method;
Multi-fractal analysis and box counting & Comparison between natural
and artificial languages \\ \hline
\cite{ausloos2016quantifying} & 
Rank-Size analysis; Zipf's law; Exponential law; Shannon entropy &
Analysis of referees responses in a peer review system\\ \hline
\cite{calude2011we} & Frequencies analysis of words from Swadesh
fundamental vocabulary; Principal component analysis; Linear
regression model; Bayesian Markov chain Monte Carlo model; & Common
words usages in different languages\\ \hline \cite{defortuny} &
Frequencies analysis of words belonging to different sentiment
classes; Knowledge discovery in databases; Polarity analysis &
Sentiment analysis of political related articles\\ \hline
\cite{ferrer2010random} & Random texts generation; Zipf's law;
Parametric distance measures; Statistical tests & Comparison between
real and random texts\\ \hline \cite{FEUERRIEGEL201888} &
Frequencies analysis of words belonging to different classes;
Extreme gradient boosting; Principal component analysis and
regression; Random forest; LASSO and Ridge regressions; Elastic Net
& Text mining of regulatory disclosures for finance applications\\
\hline \cite{griffiths2004finding} & Latent Dirichlet Allocation;
Markov chain Monte Carlo and heat bath algorithms  & Topic modeling
for PNAS papers abstracts\\ \hline \cite{herdan1958language,
herdan1966advanced} & Literature and methodological review &
Statistical linguistic exploration using many languages\\ \hline
\cite{i2003least} & Information theory; Zipf's law; Shannon entropy
 & Questioning language evolution and Zipf's law principle\\ \hline
\cite{KRAUS201965} & Discourse Parsing
(\cite{ji2014representation}); Rhetorical structure theory; Long
short-term memory; Deep Learning and Neural Network & A
methodological proposal in sentiment analysis \\ \hline
\cite{lim2002five} & Frequencies analysis of words belonging to
different classes; General Inquirer; Key words analysis & Content
analysis of the main US Presidents speeches\\ \hline
\cite{milizia2014out} & Frequencies analysis of words belonging to
different classes; Custom words classes based on geography and
political criterion & Sentiment analysis of UK political leaders\\
\hline \cite{moro} & Frequencies analysis of words belonging to
different classes; Latent Dirichlet allocation & Research of trends
in economics literature \\ \hline \cite{mostafa} & Frequencies
analysis of words belonging to different classes; Lexicon based
method; QDA Miner 4.0 software package; Multidimensional scaling &
Sentiment analysis for consumer behaviors\\ \hline
\cite{nassirtoussi2014text} & Literature review & Sentiment analysis
for finance application\\ \hline \cite{noh} & Text analyst 2.1
software; Clustering; K-means; ANOVA analysis & Keyword-based patent
analysis\\ \hline \cite{oberreuter2013text} & Uni-grams analysis;
Texts representation through vectors of frequencies; Text distance
measurement; Clustering & Intrinsic plagiarism detection model \\
\hline \cite{piantadosi2014zipf} & Literature review & Rank-size and
Zipf's law in natural language\\ \hline \cite{popescu2009word} &
Literature and methodological review & Words frequencies
applications \\ \hline \cite{saravanan2006probabilistic} & Term
frequency - Inverse document frequency; K-mixture model &
Multi-document summarizer proposal \\ \hline
\cite{saravanan2008automatic} & Term distribution model; Conditional
random field; Rule-learning system SLIPPER & Rhetorical role
identification from legal documents \\ \hline \cite{SIDOROV2014853}
& syntactic n-grams analysis; Part of speech tags; Support vector
machine; Naive Bayes; J48 classifier; WEKA approach, see
\cite{hall2009weka} & Authorship attribution\\ \hline
\cite{stamatatos2009survey} & Literature review & Authorship
attribution problems\\ \hline \cite{tang2013recognition} & Power-law
analysis of words frequencies; Frequency Rank Ratio (FRR); Influence
Language model & Analysis of words collocation within texts  \\
\hline \cite{tetlock2007giving,tetlock2008more} & Frequencies
analysis of words belonging to different classes; General Inquirer;
Principal component analysis; Regression analysis; & Sentiment
analysis for finance application \\ \hline
\cite{vejdemo2016semantic} & Frequencies analysis of words belonging
to different classes; Regression analysis  & Lexical replacement
analysis in different languages \\ \hline \cite{westergaard} & Named
Entity Recognition; Dictionary based analysis; Regression analysis &
Text mining of biomedical literature\\ \hline \cite{zeng2012topics}
& Term distributions analysis; Zipf's law; Latent Dirichlet
Allocation & Topic modeling with the Zipf's law contribution\\
\hline
\end{tabular}
\caption{List of references about text mining researches. Each row
represents a study, starting from the left side, in the first column
there are the references, in the second the approaches used and in
the third the fields of application of the respective papers.}
\label{textminingtable}
\end{table}

A key step of this research is the text-mining procedure for the
creation of the dataset which contains about 1000 Presidents'
speeches (see e.g. \cite{defortuny} as reference of a similar way
for collecting data from internet). Rough data has been taken from
the website: \url{http://millercenter.org} the $30^{th}$ of July,
2017, i.e. a set of 978 speeches, ranging from the \textit{Inaugural
address} of George Washington (1789) to the Donald Trump's speech
\textit{ Address to Joint Session of Congress} (2017) (as explained
in the Section \ref{Dataset building}, the number of speeches is
reduced to 951 following data collection and treatment phases).

Two classes of results are derived: the first one is associated to
the dataset and the second class of findings relies on the rank-size
analysis, whose parameters have peculiar meanings. The former class
comes from the building procedure presented in a phase-wise
including the pre-process phases, to ensure a comfortable
replicability of it in other contexts and for further studies. In
the respect of the latter case, we get non-linear regression on the
Zipf-Mandelbrot law and the respective goodness of fit, hence
arguing a common macro-structure among the speeches of the US
Presidents.

To the best of our knowledge, this paper is the first one dealing
with the US Presidential speeches with a so large dataset.

Interesting implications of social, political and linguistic nature
on the speeches delivered by the US Presidents can be carried out.
In fact, some noticeable regularities arise. First of all, the
number of words used in the speeches form two different regimes in
the periods 1800-1850 and 1850-1900. Speeches are more scattered --
in terms of the number of employed words -- in the last period of
investigation. However, after 1900, the frequencies of the words
seem to converge toward more homogeneous distributions. Furthermore,
a decreasing trend in the highest relative frequencies of the
speeches can be observed. All the speeches are associated to
best-fit curves with similar decays from the high to the low ranks.
All in all, results suggest the presence of common behaviors of the
Presidents when delivering a talk -- sometimes, associated to
specific historical periods -- hence pointing to waves of imitative
behaviors and institutional universal channels of communication.

The rest of the paper is organized as follows. Next section provides
the details of the method used. In Section \ref{Dataset building} a
description of the dataset building procedure is reported. Then we
propose a description of the collected data. Next, we show the main
features of the employed rank-size analysis. In Section
\ref{Result}, the results of the analysis along with a discussion of
them are reported. Finally, some conclusive remarks are presented
and future research directions are also traced.

\section{Methodology}
\label{methods}
The present study is performed through a rank-size analysis on a
large corpus constituted by US Presidents speeches transcripts. The
size is  defined as the (absolute or relative) words' frequencies in
each discourse, while the words rank is its position in the
decreasing sorted list, so that rank 1 is that for the most often
pronounced word in a speech. The  rank-size analysis technique is a
well-recognized method to explore the property of a large set of
data when the data spans several decades and when error  bars are
not precisely defined due to sampling conditions.

The power law and Pareto distribution with unitary coefficient were
originally introduced in \cite{zipf1935psycho,zipf1949human} to
explore the rank-size relationship in the field of linguistics.
After its debut, several contributions have investigated the
properties of the power law. We report some examples:
\cite{ioannides2003zipf,gabaix2004evolution,
dimitrova2015primacy,cerqueti2015evidence} in the context of
economic geography; \cite{axtell2001zipf,yoshi2004zipf,bottazzi2015zipf} in the business size field;
\cite{li2002zipf} in biology; \cite{levene2001zipf,
maillart2008empirical}  in informatics;
\cite{manaris2005zipf,zanette2006zipf}, in the context of music;
\cite{huang2008investigation} in the context of fraud detection;
\cite{blasius2009zipf} in the gaming field. For a wide review of
rank-size analysis see \cite{pinto2012review}. See Table
\ref{ranksizeusagesstable} for a summary of the just mentioned
fields of applications.

For what concerns the specific case of text analysis, the rank-size
method has been applied in a very wide range of cases, see e.g.
\cite{montemurro2001beyond,piantadosi2014zipf,
ausloos2008equilibrium, ausloos2010punctuation,
i2003least,ausloos2016quantifying,tang2013recognition}. The
approach with Zipf's law for text analysis has been criticized by
some scientists (see \cite{herdan1958language,herdan1966advanced})
but, recently, \cite{ferrer2010random,popescu2009word} had provided
clarifications on some controversial aspects.

The law has been extended by many researchers by introducing
methodological modifications as in \cite{popescu2010zipf}, or in the
Zipf-Mandelbrot law (ZML, hereafter) presented in
\cite{mandelbrot1953informational,mandelbrot1961theory} and
\cite{fairthorne2005empirical} and in the Lavalette law (see
\cite{lavalette1996facteur}), which have been proven to provide a
spectacular fit of rank-size relations, even when Zipf law fails to
do it (see e.g. \cite{cerqueti2015evidence,ausloos2016universal}).
In Table \ref{methodstable} there is a summary of the methodological
works here reported.

These references constitute a non exhaustive list of the wide number
of contributions on this topic. Indeed, the families of
laws proposed after the first Zipf's paper has massively grown. The
researchers have tailored solutions for exploring dataset that, due
to their peculiarities, leads to modification of the power laws.
Part of this phenomenon originate from the lack of a unique
theoretical model that can justify the origination of rank-size
regularities. Consequently, this paper moves from the preliminary
decision of the law for assessing the rank-size relation among many
functions in the literature. In doing so, we have restricted the
selection to researches grounded on a similar approach and stemming
from analogous theoretical premises, finding e.g.
\cite{i2006language,i2005consequences,popescu2009word}. Finally,
we follow the steps of \cite{popescu2009word} (Chapter 9) in which
the authors have evaluated the performance of different comparable
rank-size functions, which are suitable even in our cases.
Differently from the quoted paper, we do not study texts in
different languages and focus only on English texts. Therefore, the
best function for these authors -- a truncated Zeta distribution -- is not
the best one for us, because of the different valuation criteria.

In our study, the Presidents speeches transcripts are investigated
as a unique expert system of communication. Hence, the corpus is analyzed as a whole framework through a set of parameters able to capture the regularities and the patterns of the communication system. Such patterns are characterized basically by three elements:
the words frequencies at high ranks, the changes in occurrences from
high to low ranks and the weights of the tail (at low ranks) for
each speech. A detailed explanation of these elements is provided in
the following sections.
This said, it is clear that three parameters are needed to reach
such an objective. Indeed, three parameters effectively summarize
the state of the communication scheme at speech level and,
furthermore, are appropriate for representing the system in a three
dimensional space where each dimension corresponds to an
aforementioned feature. We notice that we do not fully explore here
the potentiality of such a representation. However, we recognize the
importance of this point for making differences with respect to the
bag-of-word, count vector or term frequency-inverse document
frequency (TF-IDF) approaches (e.g. \cite{tang2013recognition}) and
with respect the PCA and LSA techniques for dimensional reductions
(e.g. \cite{FEUERRIEGEL201888}).
Given these premises, we exclude all the functions with less than
three parameters.

The decision making process to chose among the rank-size laws
requires a further cost-benefit valuation. Indeed, it is necessary
to take into consideration the potential bias coming from the
estimations of three or more parameters in a non-linear relationship
due to the risk of falling in local minima. In particular, the
selection of laws dependent on a large number of parameters is
apparently more informative in its final result. In fact,
theoretically, the presence of a large number of parameters to be
calibrated leads to a wider set where identifying the best-fit
curve. Despite this mathematical evidence, the computational
experience leads to a different reasoning. Indeed, a large number of
parameters has the severe flaw of the probable inconsistency of the
obtained calibrated parameters, which are noticeably affected by the
choice of the initial point of the calibration exercise. Thus, local
minima  of a given error function are likely found, and the
computational complexity of a regression procedure with a large
number of parameters leads to the practical impossibility of
identifying the global minimum, hence leading to an estimation bias
(see e.g. \cite{ausloos2016universal} for discussion on this point).

Moreover, literature reports rare cases where the ZML has been
criticized for the estimation bias problem. As already said, the
main criticisms are grounded on the debate on the theoretical ground
of any rank-size regularity (see e.g. the striking cases of
\cite{herdan1958language,herdan1966advanced}). Concluding, three
parameters are enough for describing the system we are dealing with
and do not lead to bias estimation problems. Hence, Zipf-Mandelbrot
law, which has also been commonly used in this field, is appropriate
for our analysis.

We conclude this methodological section with some technical details.
We implement the best fit procedure by using the Levenberg-Marquardt
Non-linear Least-Squares Algorithm in order to derive the ZML
parameters for each speech (in Section \ref{RSZML} the details are reported). In so doing, we obtain a collection of
951 triples of best-fit parameters on the absolute frequencies as
much as for the analysis over the relative frequencies (see the next
section for a justification of the speeches number reduction).

\begin{table}
\centering
\resizebox{\textwidth}{!}{\begin{tabular}{M{0.2\textwidth} |M{0.4\textwidth} | M{0.4\textwidth}}
\hline
    \textbf{Authors}  & \textbf{Model} & \textbf{Description}\\
\hline \cite{axtell2001zipf} & Zipf's law & Analysis of distribution
of U.S. Firms; Economics\\ \hline \cite{blasius2009zipf} & Zipf's
law; & Analysis of chess opening distribution; Gaming \\ \hline
\cite{bottazzi2015zipf} & Zipf's law; Hill estimator; Literature
review & Critical review of firm size distribution studies;
Economics\\ \hline \cite{cerqueti2015evidence} & Zipf,
Zipf-Mandelbrot and Lavalette's law & Italian cities tax income
distribution analysis; Economic Geography\\ \hline
\cite{dimitrova2015primacy} & Zipf's law; Length ratio & Bulgarian
Urban system investigation across years; Economic Geography\\ \hline
\cite{gabaix2004evolution} & Zipf's law; Gibrat's law; Hill
estimator & Urban growth and City size distribution analysis;
Economic Geography\\ \hline \cite{huang2008investigation} & Zipf's
law; Benford's law & Proposal of a fraud detection model\\ \hline
\cite{ioannides2003zipf} & Zipf's law; Gibrat's law & Urban growth
and City size distribution analysis for the US case; Economic
Geography\\ \hline \cite{levene2001zipf} & Zipf's law; Markov chain
& Web surfer behavior modeling; Informatics\\ \hline\
\cite{li2002zipf} & Zipf's law & Genes distribution; Biology\\
\hline \cite{maillart2008empirical} & Zipf's law & Linux
distributions diffusion analysis; Informatics\\ \hline
\cite{manaris2005zipf} & Zipf's law; Zipf-Mandelbrot law &
Authorship attribution model; Music \\ \hline \cite{pinto2012review}
& Power law; & Applications review
\\ \hline \cite{yoshi2004zipf} & Zipf's law & Japanese
firms bankruptcy distribution analysis; Economics\\ \hline
\cite{zanette2006zipf} & Zipf's law; Simon's model &
Connection between linguistic and music investigation; Music \\
\hline
\end{tabular}}
\caption{List of references of different rank-size applications
(excluding text mining). Starting from the left side, the first
column reports the references of the papers, the second one lists
the employed models and the third column contains a small summary
and the scientific fields in which the studies are collocated.}
\label{ranksizeusagesstable}
\end{table}

\begin{table}
\centering
\resizebox{\textwidth}{!}{\begin{tabular}{M{0.2\textwidth} |M{0.4\textwidth} | M{0.4\textwidth}}
\hline
    \textbf{Authors}  &  \textbf{Approach} & \textbf{Description}\\
\hline \cite{ausloos2016universal} & Zipf, Zipf-Mandelbrot and
Lavalette laws & Proposal of a new, more general, rank-size law\\
\hline \cite{fairthorne2005empirical} & Bradford-Zipf-Mandelbrot
laws & Literature review of the empirical hyperbolic distributions
\\ \hline
\cite{i2005consequences} & Zipf's law & Signal model for modeling
communication considering syntax and semantic and new proposal\\
\hline \cite{i2006language} & Zipf's law & Signal model for modeling
communication and new proposal \\ \hline \cite{lavalette1996facteur}
& Lavalette's law & Paper from Lavalette stating his famous law\\
\hline \cite{mandelbrot1953informational, mandelbrot1961theory} &
Zipf-Mandelbrot law & Papers from Mandelbrot proposing the
modification of the famous Zipf's law\\ \hline
\cite{popescu2010zipf} & Rank-size laws & Proposal of a model with
different rank-size regimes within Zipf-like distributed data\\
\hline \cite{zipf1935psycho, zipf1949human} & Zipf's law & Papers
from Zipf stating his famous law\\ \hline
\end{tabular}}
\caption{List of relevant references included in the present paper
associated to methodological proposal, survey or text mining
rank-size techniques. Staring from the left side, in the first
column one can find the references of the papers; the second one
contains the discussed methodological tool(s) and the last column
summarizes the nature of the contributions.} \label{methodstable}
\end{table}

\section{Dataset building}
\label{Dataset building} This section is devoted to describe the
dataset building process, namely we  present the routine's actions
imposed by using R. The commands used to build the dataset are
provided by the libraries \textit{"xml2","rvest", "stringr",
"xlsxjars", "xlsx"} along with their respective dependencies. From a
text mining point of view, a part for the web-scraping phases, the
following steps are in line with \cite{tang2013recognition,oberreuter2013text}.

The building procedure is divided in 13 phases.

In the first step, the considered
website was visually examined in order to understand the structure of the contents. In
particular, since one is looking for the presidential talks transcripts, it is
important to find the addresses' pages where they are listed. In the
case of the Miller Center web site, the hyperlinks to each speech are
dynamically showed in the following page: \url{https://millercenter.org/the-presidency/presidential-speeches}.
Consequently, one has to inspect the HTML source code for finding
the objects of interest and to decide how to select them. In this
case, it means that one needs to save the whole source HTML code and
extract  the located links as showed in the next phase.

The second step is devoted to the first moves of the \textit{web
scraping technique} as described even in \cite{jockers2014tex,munzert2014automated} for the specific case of R. Such a technique
is an automated computer science procedure for extracting
information from websites through a combination of dedicated
commands. Thanks to them, it is possible to systematically access
web pages in order to extract data of interest. The phases of a web
scraping routine are twofolds: the first stage is characterized by
the saving of the HTML source code
of the web pages; the second stage consists of
the extraction of the portion of the code
where the needed information is reported.
These actions are performed thanks to the functions:
$read\_html(),html\_nodes(),html\_attr()$ and $html\_text()$.
Such functions are available in the "rvest" library,
which is employed for systematically
finding  the links to speeches' transcripts and other contents of interest.\\
So, in the second step, one grabs all the URLs (the acronym standing
for: Uniform Resource Locators) of the speeches in order to prepare
a list of links to be opened. The process of saving the addresses
from the reference web page might lead to the occurrence of some
errors. Such errors can be including mistakes produced by the web
site creator. As an example, the links to the speeches' pages could
be reported by a different HTML identifier into the page, and this
would lead to empty memorization. A control procedure has been
applied in order to face this problem;  at the end of this phase,
978 web pages' addresses to the speeches' transcripts were obtained.

The third step consists in the application of another web scraping
routine on each page that contains the transcribed words. The list
of links pre-loaded in the previous phase is treated to obtain
speeches' transcripts, titles, dates, places of the statement,
sources, and resumes of the speeches. This step is implemented
through a \textit{"for"} loop over the list of links that point to
the pages where the speeches are stored. In each $for$'s cycle, the
web scraping routine is applied for the second time in order to read
the HTML code of each web page pointed by the $for$ running index.
At the same time, one controls for possible discrepancies that could
occur through an \textit{"if"}'s statements inside the loop. This
has to be done because sometimes the web pages where the speeches
are presented could contain errors like: blank area where the corpus
is supposed to be and/or the transcript is reported into another web
page's section. 
In so doing we have realized that the web page that contains the
speech: "Campaign speech in Indianapolis, Indiana" stated by Herbert
Hoover in October 28, 1932
\cite{indianapolis},
is one of the remarkable exceptions. Indeed, in the corresponding
web page the discourse's transcript is positioned in the section
dedicated to the resume. So, to capture it, we have used the same
HTML selector used for memorizing the resumes of the talk, which is
usually positioned in the top-right side of each web page.

At this stage, a visual inspection of the obtained results
highlights that different links in the list point to the same
speeches' pages. Consequently, in the third step, some information
would be downloaded twice. In order to manage such a case, a control
on the speeches' title has been applied. In particular, we have
checked for titles duplication and saved their position. In this
way, it is possible to eliminate the respective positions into the
variables used for each type of information (titles, resumes etc..).
There are 7 so obtained duplicated speeches: January 20,
2005: "Second Inaugural Address", April 27, 1961: "President and the
Press",  June 12, 1895: "Declaration of US Neutrality", December 6,
1892: "Fourth Annual Message", December 9, 1891: "Third Annual
Message", December 1, 1890: "Second Annual Message" and December 3,
1889: "First Annual Message". After this further control, we have
971 stored transcripts.

The fourth phase is employed to manage the presence of typos in the
inspected web pages of the Miller Center web site. The typologies of
typing errors that mind more for the analysis are all those that
contrast a correct division of the text into different tokens.
Examples are the situations where the space between two words,
number and word or punctuation and word is missing. Such typos,
which can generate strings like: \textit{"you.Therefore"},
\textit{"10,000of"} or \textit{"thePresident"} etc., impede the
software to divide
the text according to the adopted tokenization method.\\
The procedure for managing such typos is the following: the
transcripts are firstly stored as a list of strings in a variable.
One looks into each string to find all words divided by points
without spaces like: " years.And ", " slowed.And ". These two
elements  are for example found in: "2016 State of the Union
Address" \citep{obamasou2016}.
The problem is solved by inserting a space between the points and
the following word. 
\\
Then, one solves the issue of  numbers followed and preceded by
words without spaces' interruptions.
An example of this typo can be found in the "First Annual Message" of December 6,
1981 \citep{chesterfam1981}.
There, the following exceptions occur: "June30", "in1881", "length3", "since1860", "of250".
With the same method used in the previous phase, this typos are found and corrected as follow: "June 30", "in 1881", "length 3", "since 1860", "of 250".\\
Lastly, one manages the typos generated by two consecutive words
merged without spaces, with the former entirely made by lower case
characters and  the latter made by the first character in upper case
and the rest in lower case. An example is in the "Inaugural Address"
of March 4,
1925 \citep{coolidgeia1925},
where the wrong token " ourConstitution" has been transformed through
the correction
process in " our Constitution".\\
The procedure described in the fourth step is developed by employing
regular expressions, which is a simplified method for searching
patterns into strings by means of pseudo-coding languages. A
formal definition of the regular expression is reported in \cite{mitkov2005oxford}.

A further problem is related to the interactions of the President.
Indeed, the President often talks in front of a wide and active
audience. In such cases, it could happen that he is interrupted by
applause, laughters, singed slogans or very loud screams at which
the President could sometimes respond. These situations are reported
into the speeches transcripts: the applause and laughter are
sometimes reported between round or square brackets, while other
kinds of contents like interactions between Presidents and audience
are displayed after the specification of the speaker. A quite
complete example of the described situations is given by the speech
stated by Barak Obama: "Remarks in Eulogy for the Honorable Reverend
Clementa Pickney" \citep{obamarem2015}.

Thus, the fifth phase is devoted to remove the parentheses (square
and rounded brackets) and their contents. For  this aim, one needs
to systematically access the individual speeches. In so doing, for
example, there are cases in which the parentheses appear in  the
text but as typos; see e.g. the "Sixth Annual Message" of December
4, 1928 \citep{coolidgesam1928}. This type of exceptional error can
be detected by registering the lengths of the parentheses' contents.
For the analyzed dataset, the non-suspicious length of parentheses'
content amounts to about 600 characters. In order to identify this
limit as reasonable for
  a string length  between two brackets, a visual inspection of all
the parentheses' contents has been performed. Above the critical
threshold of 600 characters, one can consider that the strings
bounded by two parentheses do not constitute a content to be
eliminated but, rather, a typo (this is the case of a missing
closing bracket). A control is implemented by means of an
\textit{"if"} statement, to check if the eliminated pieces of text
do not exceed the
  600 characters threshold.

In this context, we need to mention also public events like press
conferences, that are typically followed by questions from the
public or by the journalists. The questions are denoted by an
initial uppercase "Q" followed by a punctuation character or a
space.\\
Another well established type of Presidential public meeting
are the debates, which are characterized by a dialogue among
candidates and/or journalists.
\\
The above-mentioned elements constitute noises for the analysis of
the Presidential statements;  therefore, the sixth phase is
dedicated to the amendment of the transcripts from the strings that
do not come out directly from the Presidents, unique source of
interest of this study. In this phase, some speeches with peculiar
complexity (like the debates, where the rhetoric structure of the
Presidential speech may be questionable, being also driven by the
conversation flow) have been directly removed.
In particular, we have removed 13 Debates and 1 Conversation
from the original Miller Center database. To this end, we have
looked for the words "Debate" and "Conversation" into the titles list and removed all the transcripts corresponding to the titles in which such
strings appear.
\\
By means of a \textit{"for"} loop, the transcripts have
been fathomed in order to find the presence of "Q" followed by
punctuation or space, because such strings are clear signals of the
presence of a question (so, out of the President rhetoric, as
mentioned above). For consistency, all the text after a question has
been removed; indeed, Presidents words are driven by the
conversation and are not relevant in the analysis of the rhetoric
structure of the speeches. An example is the
Ronald Reagan speech: "Speech on Foreign Policy" state in December
16,
1988 \citep{reganfp1988}.
Some exceptions did arise. For example "Q." is often used into the
transcripts to report names' abbreviations, like \textit{"J. Q.
Adams"} or \textit{"Q. Tilson"}. Such cases have been treated
through a visual inspection of the removed texts.
\\
Furthermore, in the Miller Center speeches' transcripts structure,
the most noticeable words pronounced by the public and reported are
preceded by strings like: "THE AUDIENCE", "AUDIENCE", "AUDIENCE
MEMBER" or "Audience", while the President statements are preceded
by strings like: "THE PRESIDENT" or "President" sometimes followed
by the President's surname (i.e. see the speech "Remarks at the
Democratic National Convention" state by Bill Clinton in August 29,
1996 \citep{clintonrem1996}).\\
By using the regular expressions, it is possible to remove the
reported public interventions, thus leaving just the words
pronounced by the President. Such an amendment is done by
eliminating the characters between the second (or third in case  there were
 multiple speakers like auditors and journalists) speaker markers
(for example "AUDIENCE") and the President markers (for example "THE
PRESIDENT"). This means that one has to meet jointly (one after the
other) the markers of both speakers in order to select the
unnecessary text portions. Consequently, this process might lead to
the removal of an excessive piece of text in the unlucky case of an
error in the transcript. For example, this is the case of the
presence of a
second speaker marker that is not followed by the President's marker
once he takes the floor again, and the missing marker appears only
later in the text. To manage such error cases, we proceed by further
analyzing the texts candidate to be removed. In particular, we
consider a threshold of 400 characters above which the selected
texts are not canceled. That is decided thanks to a visual
inspection of all the portions of the selected texts. A remarkable
example of the described exception is the speech
 by George W. Bush on September 3, 2004, entitled: "Remarks at
the Republican National
Convention" \citep{gwbushrem2004}.
In that transcript, there are many audience interventions which are  reported
and marked with the string "AUDIENCE:" but after one of them there
is a lack of  marker that should have  indicated where the President'
words appear again. Thus one would eliminate a bigger portion of
text that ends when the string "THE PRESIDENT:" is met again. In
order to avoid the loss of large bunch of data, we decided to leave
the selected pieces of text longer that 400 characters inside the
analyzed sample. After all the included words are so few that they cannot
affect the final result; this  was  confirmed by visually inspecting
the few exceptions which were  found.\\
Sometimes, at the very  beginning of the speech transcript, it  is
possible to find a string of the type: "THE PRESIDENT:", "The
President:" or "The President" followed by his surname to mark the
starting point of the President's words. This string is not captured
by the described process because it is the very first and is not
coming after the intervention of other speakers. Therefore, at the
end of this phase, one has to control for the presence of strings of
that type at the opening words of the text, and has to eliminate
them in affirmative case.

The seventh phase manages the situations in which the President
delivered messages with his wife. In this case, we have applied a
control using a string of the type \textit{"Mrs. "} followed by the
President's surname. Then, to meet consistency, we have removed all
the words of the speech from the starting point of the intervention
of President's wife.

At this stage, one has to check that the listed modifications do not
reduce a given speech so much that it becomes not suitable for
performing a consistent analysis. For exploring the implications of
too short corpus in applying ZML fit see \cite{dkebowski2002zipf}.
The control of the suitable length of the speeches is the scope of
the eighth phase. We have eliminated the speeches whose resulting
number of characters is less than 600. This threshold has been
identified by the inspection of the distribution of the number of
characters for each speech. After this procedure, we have eliminated
five speeches: "Press Conference with Mikhail Gorbachev" (July 31,
1991), "Press Conference" (November 17, 1967), "Press Conference"
(August 18, 1967), "Press Conference" (December 30, 1966), "Argument
before the Supreme Court in the Case of United States v. Cinque"
(February 24, 1841).

A further control for identifying speeches with the intervention of
people different from the President has been next implemented. In
particular, in some Press Conferences the question are introduced by
the name of the journalist followed by the name of the newspaper. In
order to capture such cases, one controls for the presence of the
string "Press Conference" in the titles and for the absence of the
mentioned markers of the questions ("Q" followed by blank or
punctuation). This type of exception is met just with: "Press
Conference in the East Room", July 20, 1966
\citep{lyndonpresconf1966}. This speech has been eliminated from the
list. In fact, from a visual inspection, one can see that it is
mainly made of words not provided by the President but rather by
journalists.

The ninth phase is devoted to a last check of the outcomes of the
previous steps. In particular, the string: "THE PRESIDENT" followed
by punctuation or by the surname of the President (all upper or
lower case) has been seek in the speeches. The presence of such a
string points to talks in which there is the intervention of the
public or other speakers; such interventions did not appear in the
previous phases since they are placed in the first line of the
speeches. As an example, refer to the Obama's speech: "Address to
the United Nations" delivered in September 23,
2010 \citep{obamaaun2010}.
In   other cases, the string is reported at the beginning of the
speech just for indicating the point in which the president is
starting to speak. This control  was  run over the entire dataset
and not just on the modified transcripts as for the previous phases.

The tenth stage is called \emph{tokenization
phase}. A formal definition and some practical examples of tokenization are reported in
\cite{manning2008introduction}.
Indeed, when the talks' transcripts are stored into the cells, long
strings of characters are memorized without any particular
distinction. Yet, in order to work with words' frequency, one needs
to refine the speeches until they become a list of comparable units of analysis. 
Consequently, one needs to split these strings of characters
(one for each speech) by making the procedure in R  to be able to recognize
single words in accord to the used tokens definition.\\
Specifically, with the R library "tokenizers", it is possible to use
the command $tokenize\_words()$. Such a command divides the text by
using the blank space as a separator and without taking into consideration
the punctuation except for the decimal and thousand numbers separators. 
Furthermore it does not consider the apostrophes between
words as a separator, like in the case of contract form of verbs.
As an outcome, we obtain that the variable containing the entire
speech is transformed into a vector whose components are the words.
Moreover, all the letters of the words are converted from upper case
to lower case. In so doing, possible ambiguities due to the case
sensitiveness of the words is forcefully removed.

\begin{figure}[h]
\centering
\includegraphics[width=5.7in]{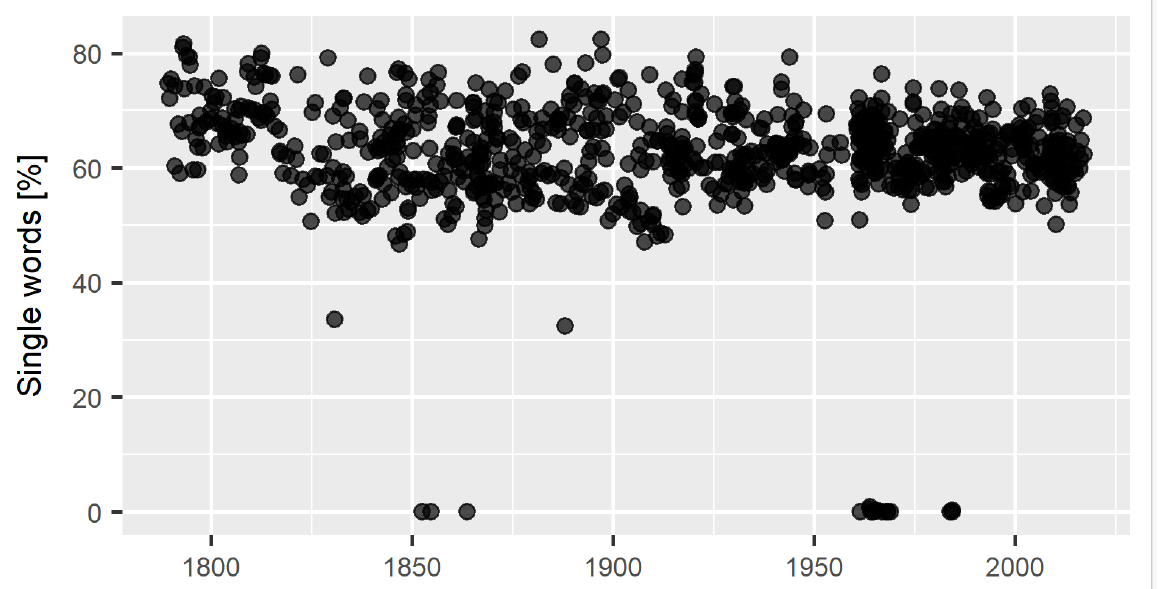}
\caption{Percentage of words used just once on the
number of different words used in each speech in a time varying representation.}
\label{Fig1}
\end{figure}

In the eleventh step we have implemented a control for the speeches
that are doubly reported  into the same page section, hence leading
to doubling the frequencies of the words. Sometimes it happens that
those repeated transcripts do not have exactly the same words in
common, but differ for few terms (for a maximum of about 20,
according to our empirical experience). This can be noticed by
observing that some speeches have very few words with frequency
equal to one, the so-called hapax legomena. For this reason, the
control for finding the double repeated transcripts is done by
checking if the number of tokens that appear only once falls below a
certain threshold.
Thus, for each speech,  we divided the number of words occurring just once, by the number  
of different words used, (see  Figure \ref{Fig1});
thanks to a visual inspection, one fixes the critical threshold at 20\%.\\
The failure of this check pointed out to some technical
inconsistencies of the website; one of the affected  transcript  is:
"Remarks Honoring the Vietnam War\'s Unknown Soldier" of May 28,
1984 \citep{regamrem1984}.
To solve this bug, we divided the terms' frequencies by 2, each time
the control on the threshold was failing. But, in so doing, the
exceptional terms appearing once in the double reported speeches
reached absolute frequencies equal to 0.5. For this reason one has
to add a further control for eliminating the residual tokens with
0.5 absolute frequencies.

The twelfth phase concerns the creation of a table type variable
that has a number of rows equal to the number of different words
used in each single speech and two columns: one for the tokens and
another for the frequencies. Each couple is sorted out  by
decreasing order of frequencies. The tables of the speeches have
been labeled by the title and date of the speech.

To make the dataset exportable and ready to be processed, one
collects and save the data into a comma-separated values file (i.e.
a $csv$ file).
A matrix with $951 \times 2$ columns and 3933 rows is
then obtained. Each couple of columns is dedicated to the so sorted
list of words and their respective frequencies for the individual
speeches. In order to obtain a rectangular matrix, the number of
rows is uniformed at the maximum number of different
words used across speeches, which is 3931. Of course many speeches
have a lower number of different words, and the empty cells are
filled by \emph{NAs}, which point to a missing value indicator.
After this, one adds two rows at the top of the table: the first one
is used to report the talks' speaker name; the second one is adopted
to show the titles with the dates embedded.

As already preannounced above, the last phase concerns the export of
that table into a $.csv$ file. In so doing, one has a dataset which
is easily analyzable, -- also with different programming languages;
  this goes in the direction of making  this study reproducible.

\begin{table}[h]
\centering
\begin{tabular}{cccc}
  \hline
 \textbf{President} & \textbf{No.} &  \textbf{President} & \textbf{No.} \\
  \hline
      Lyndon B. Johnson     &66&       Jimmy Carter&    18  \\
    Ronald Reagan & 57  &   John Tyler  &18 \\
  Barack Obama  &50&       Warren G. Harding    &18 \\
Franklin D. Roosevelt   &49&        Rutherford B. Hayes     &16 \\
     John F. Kennedy    &41&    Abraham Lincoln     &15 \\
George W. Bush  &39&    Franklin Pierce     &15 \\
   Bill Clinton     &38&    Gerald Ford     &14 \\
     Woodrow Wilson     &33&    James Buchanan  &14 \\
    Ulysses S. Grant    &32&        William McKinley    &14 \\
  Andrew Johnson    &31&       Calvin Coolidge  &12 \\
Herbert Hoover  &30&         William Taft   &12 \\
Grover Cleveland    &29&    Chester A. Arthur   &11 \\
  Andrew Jackson    &26&        James Monroe    &10 \\
James K. Polk   &25&       Martin Van Buren     &10 \\
   Thomas Jefferson     &24&       John Adams   &9  \\
   Richard Nixon    &23&       John Quincy Adams    &8  \\
   James Madison    &22&       Millard Fillmore     &7  \\
 Theodore Roosevelt   &22 & Dwight D. Eisenhower  &6  \\
George Washington   &21&         Zachary Taylor     &4  \\
George H. W. Bush   &20&    Donald Trump    &2  \\
  Benjamin Harrison     &19&    James A. Garfield   &1  \\
Harry S. Truman     &19&       William Harrison     &1  \\

\hline
\end{tabular}
\caption{Number of speeches (No.) per President. The list is for
44 presidents, instead of 45, because Grover Cleveland has had two
non-consecutive mandates: the first from March 4, 1885 to March 4,
1889 and the second from March 4, 1893 to March 4, 1897. In order to
count his number of speeches, the two mandates are grouped
together.} \label{presidentbyspeech}
\end{table}
\section{Dataset Description}
\label{datasetDescription}

This section contains the description of the dataset.

At the end of the process described in the previous section, one has
a dataset of 951 Presidential speeches stated by the 45 United
States Presidents, from George Washington to Donald Trump. The
dataset covers a wide period of time: from April 30, 1789 to
February 28, 2017. Due to the Miller Center web site's content, the
number of speeches per President are different and depends on
criteria decided by the website owner. The number of the speeches
per President is reported in Table \ref{presidentbyspeech}.

The Miller Center website provides discourses stated in many
occasions of United State Political life: for example there are 57
State of the Union Addresses, 142 Annual Messages, 58 Inaugural
Addresses, 20 discourses state at universities or related to them,
18 speeches stated at National Conventions of Republicans or
Democratic parties, 89 remarks pronounced by Presidents on salient
topics and 567 other moments when the US Presidents have spoken to
people. This information is summarized in Table \ref{datacontent}.

All these declarations are collected as described in the previous section;
furthermore, they are stored by organizing the words' distribution
for each speech. In this way,  it is easy to apply the rank-size
analysis in order to investigate the different rhetoric structures,
as shown in the next section. Table \ref{summarystatisticsfreq}
presents a statistical description of the speeches' length in term
of number of words per talk (second column) and in term of the
different terms used in each speech (third column).  The statistics
will be commented upon in the result section. The discourse with the
minimum number of different words used is the \textit{"Message to
Congress Requesting War Declarations with Germany and Italy"} in
December 11, 1941, by Franklin D. Roosevelt; the one with most words
variety is the \textit{"Seventh Annual Message"} of December 3,
1907, by Theodore Roosevelt. It is interesting to note that the
impressive amount of different words used in the latter message is
related to the fact that, during this talk, the President Roosevelt
has read a part of another speech (Message to the Congress on
December 5, 1905) and has mentioned events from the past, hence
increasing the lexical richness of his speech.

\begin{table}[h]
\centering
\begin{tabular}{rrr}
  \hline
\textbf{Statistical indicators} & \textbf{Speech Length} & \textbf{Different words used} \\
  \hline
  Max. & \numprint{27551} & \numprint{3931}\\
  Min. & 132 & 76\\
  Median ($m$) & \numprint{2315} & 760\\
  Mean ($\mu$) & \numprint{3533} & \numprint{916.23} \\
  RMS & \numprint{5256.24} & \numprint{1144.87}\\
  St. Dev.($\sigma$) & \numprint{3893.80} & \numprint{686.85}  \\
  Variance & \numprint{15145734.74} & \numprint{471265.19}\\
  St. Error & \numprint{126.27} & \numprint{22.27}\\
  Skewness & 2.64 & 1.58 \\
  Kurtosis & 12 & 6.02\\
  \hline
  $\mu/\sigma$ & 0.91 & 1.33\\
  $3(\mu-m)/\sigma$ & 0.94 & 0.68 \\
   \hline
\end{tabular}
\caption{The numbers in column two offers a statistical summary of
the speeches' lengths in term of total number of words per speech.
The third column contains the statistics of the number of different
words used in each speech.} \label{summarystatisticsfreq}
\end{table}

\begin{table}[h]
\centering
\begin{tabular}{M{0.6\textwidth} M{0.4\textwidth}}
\hline
\textbf{Class of speeches} & \textbf{Number of speeches} \\
\hline
Speeches stated at National Conventions of the Republican or Democratic parties & 18 \\ \hline
Discourses stated at Universities or related to them & 20 \\ \hline
State of the Union Addresses & 57 \\ \hline
Inaugural Addresses & 58 \\ \hline
Remarks on salient topics & 89 \\ \hline
 Annual Messages & 142 \\ \hline
Others & 567\\
 \hline
\end{tabular}
\caption{The pre-processed corpus scraped form the Miller Center web site has been dived in classes which contains the number of speeches here reported.  See Section \ref{Dataset building} for the details on the data collection and manipulation procedure.}
\label{datacontent}
\end{table}

\section{Rank-Size: Zipf-Mandelbrot Law}
\label{RSZML} 
\begin{figure}[h]
\centering
\includegraphics[width=5.7in]{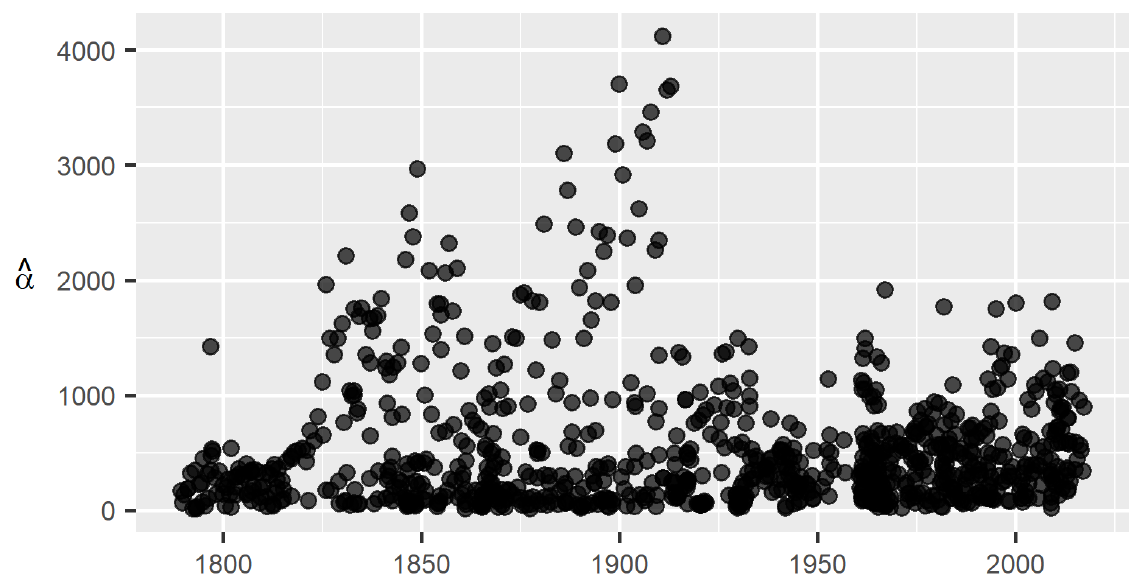}
\caption{$\hat{\alpha}$ estimated on absolute frequencies with Eq.
(\ref{ZML}) for each speech.} \label{Fig2}
\end{figure}

\begin{figure}[h]
\centering
\includegraphics[width=5.7in]{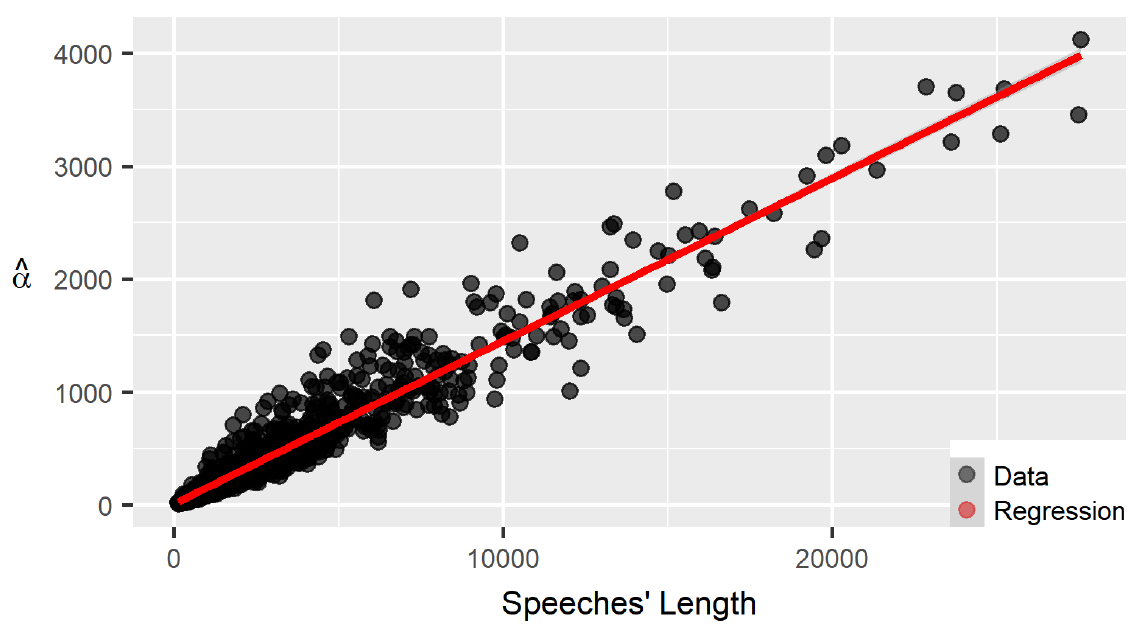}
\caption{Relationship between $\hat{\alpha}$ and speech length in
term of total number of words used per corpus} \label{Fig3}
\end{figure}

The speeches' words frequencies are  the size of the
rank-size analysis. Specifically, each talk transcript is stored
into a table with terms and respective frequencies, so the rank $r =
1$ corresponds to the most repeated word of the speech. The tokens
with lowest frequencies are stored in the positions corresponding to
the highest ranks. The use of the plural is required because the
terms with frequencies 1 and 2 represent generally the majority of
the tokens in each discourse. Here, a best fit procedure to assess
whenever the size-frequencies $f$ and $f_{rel}$ (absolute and
relative respectively) might be view  as a function of the ranks $r$
is implemented. The considered fit function is the ZML:
\begin{equation}
\label{ZML} f \sim g(r)=\alpha(r+\beta)^{-\gamma},
\end{equation}
where $\alpha$, $\beta$,$\gamma$ must be calibrated individually for
each one of the 951 speeches. All fits have been carried out through
a Levenberg-Marquardt algorithm \citep{levenberg1944method,marquardt1963algorithm,lourakis2005brief} with no restrictions on
the parameters. The starting points used in each estimation are
provided through a non linear regression run over the same function
but with a brute-force algorithm also know as grid-based searches,
to avoid the dependence on starting parameters or getting stuck in
local solutions. The same procedure is applied over the following
formula, hence on the relative frequencies of the speeches' words.
\begin{equation}
\label{ZML_rel}
f_{rel} \sim \frac{g(r)}{N}=\tilde{\alpha}(r+\beta)^{-\gamma},
\end{equation}

where $N$ is the length of the considered speech and $\tilde{\alpha}$, $\beta$, $\gamma$ are the parameters to calibrate.\\
The estimated parameters interpretations for the case of Eq.
(\ref{ZML}) are the following: $\alpha$ gives information on the
the number $N$ of words of the speech (see Figure \ref{Fig2}), which is emphasized in Figure
\ref{Fig3}, and removed in the relative frequencies case of Eq.
(\ref{ZML_rel}). This aspect is discussed more in detail in the next section.

The parameter $\beta$, contains information on the higher ranked
words. In particular the model reduces the weights of tokens with
high frequencies when $\beta$ increases. Moreover, if one does not
have the presence of outliers, $\beta$ is small.

For what concerns $\gamma$, we expect that this parameter is close
to 1, as it  will be found; see results below. This parameter
describes the concavity of the fitted models, whence is informative
about the distribution of words frequencies thereby giving an idea
about their density. Indeed, if the magnitudes of the frequencies in
the medium-high ranked words are high, then the calibrated
$\hat{\gamma}$ is  to be found small. Furthermore, $\gamma$ is
affected by the number of hapax legomena. If their presence is bulky
with respect to the rest of the tokens, the model concavity will
increase in order to capture the  transition point between high
ranks  and low ranks.

This interpretation of the parameters is coherent with
\cite{bentz2014zipf}, and  will be  also discussed  later.

\section{Results}
\label{Result} 
\begin{figure}[h]
\centering
\includegraphics[width=5.7in]{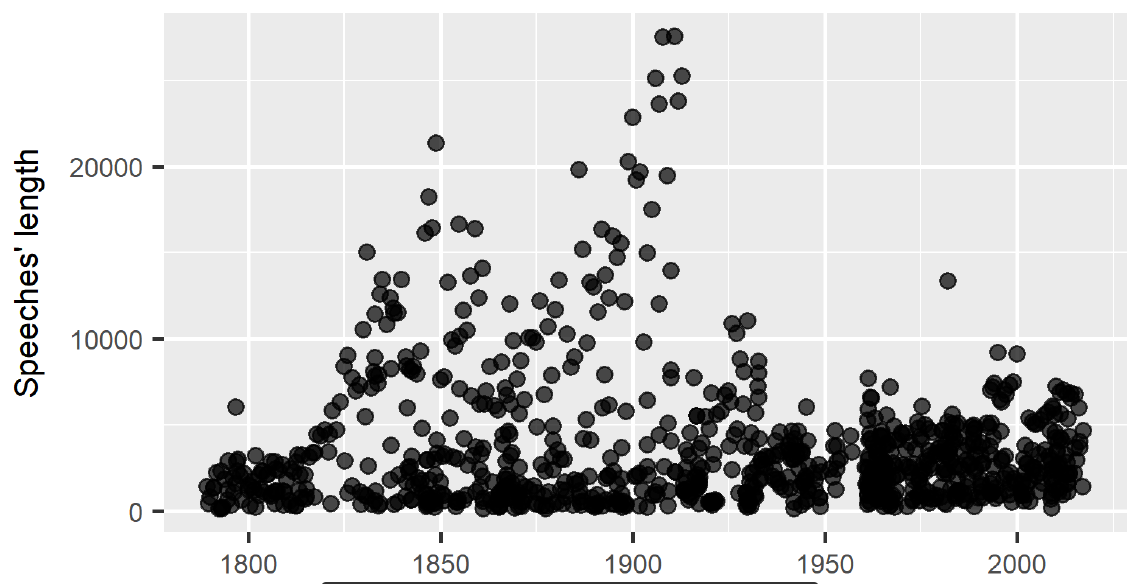}
\caption{Speeches' length distribution in term of total number of words used per speech over the years}
\label{Fig4}
\end{figure}

The second column of the Table
\ref{summarystatisticsfreq} contains the main statistical indicators
of the speeches' length in term of total used words in each speech.
When looking at such statistics we can notice that the length varies
considerably and its mean and median do not coincide. The positive
skewness suggests a right-tailed shape, and the kurtosis equal to 12
indicates a leptokurtic distribution with very heavy tails. A
similar situations is presented in the third column of Table
\ref{summarystatisticsfreq} where the main statistics of the number
of different words used in each speeches are presented. The
asymmetry is well identified by skewness value and confirmed by the
different positional indicators. As for the previous case, we obtain
a leptokurtic distribution
but with a kurtosis equal to 6.02.\\
Figure \ref{Fig4} gives an idea of the talks lengths along the
considered period of our sample and it is also informative about the
years in which we have "grader masses" of speeches by observing the
density of the points. The recent years denote a change of behavior,
indeed after a small number of speeches in the 60's, there is a
noticeable concentration of the speeches' in terms of their lengths.
In a number of cases the speeches lengths and words variety are
considerably high, due to the presence of very long speeches in
which the President is reading some other documents or is quoting
other talks.

In Figure \ref{Fig1}, it is possible to note the pattern of the
percentage of words used only once per talk. We can note also a
slightly decreasing trend that could be associated to the need of
reducing the number of different words used. In the light of the
information available here, a possible explanation of that could be
the recent communication need of repeating messages many times, with
same terms, for a simpler and more aggressive media strategy. Anyway
the speeches are mostly characterized by the presence of words
pronounced only once. Such words populate the tails of each ZML
distributions and provide a characterization of them.

\begin{figure}[h]
\centering
\includegraphics[width=5.7in]{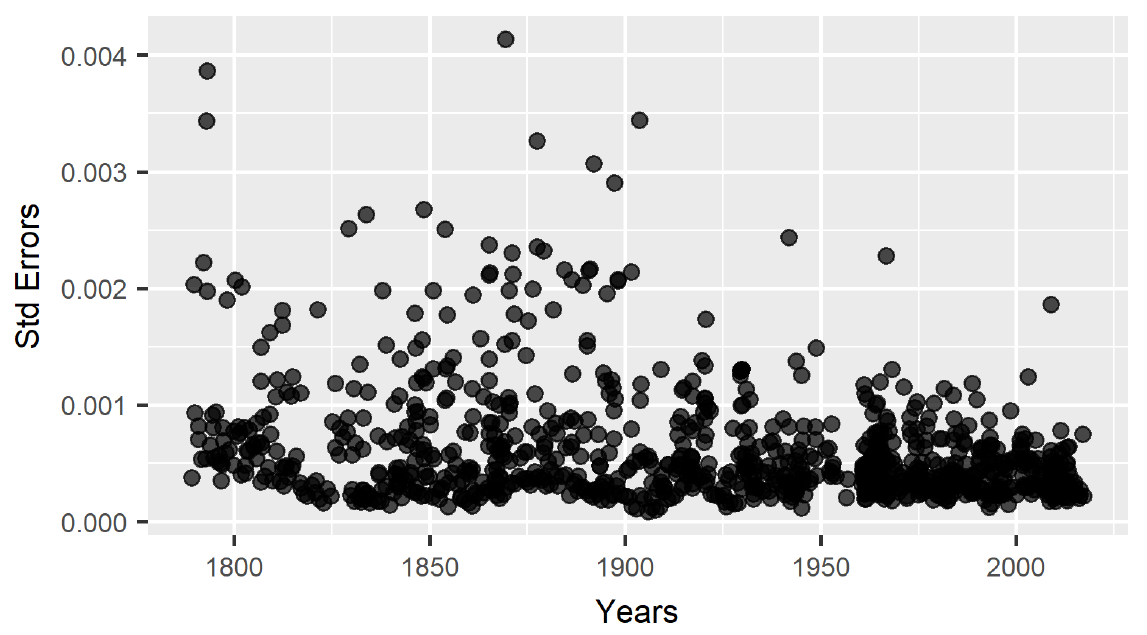}
\caption{ Regressions standard errors for each fitted speech on Eq.
(\ref{ZML_rel}) over the years.} \label{Fig5}
\end{figure}

The best fit procedure on Eq. (\ref{ZML}) and Eq. (\ref{ZML_rel}) is
performed for any individual speech, and a visual presentation of
the goodness of fit measure for the second equation is reported in
Figure \ref{Fig5}. The main stats of $R^2$s and regression standard
errors for both equations are presented in Table
\ref{summarystatisticrsq}. Furthermore we test the normality of the
standardized regressions residuals by using the Shapiro--Wilk test
(see \cite{shapiro1965analysis}); the p-values are reported in Table
\ref{summarypvalues}.

From the analysis of the absolute frequencies, some facts emerge.

\begin{figure}[h]
\centering
\includegraphics[width=5.7in]{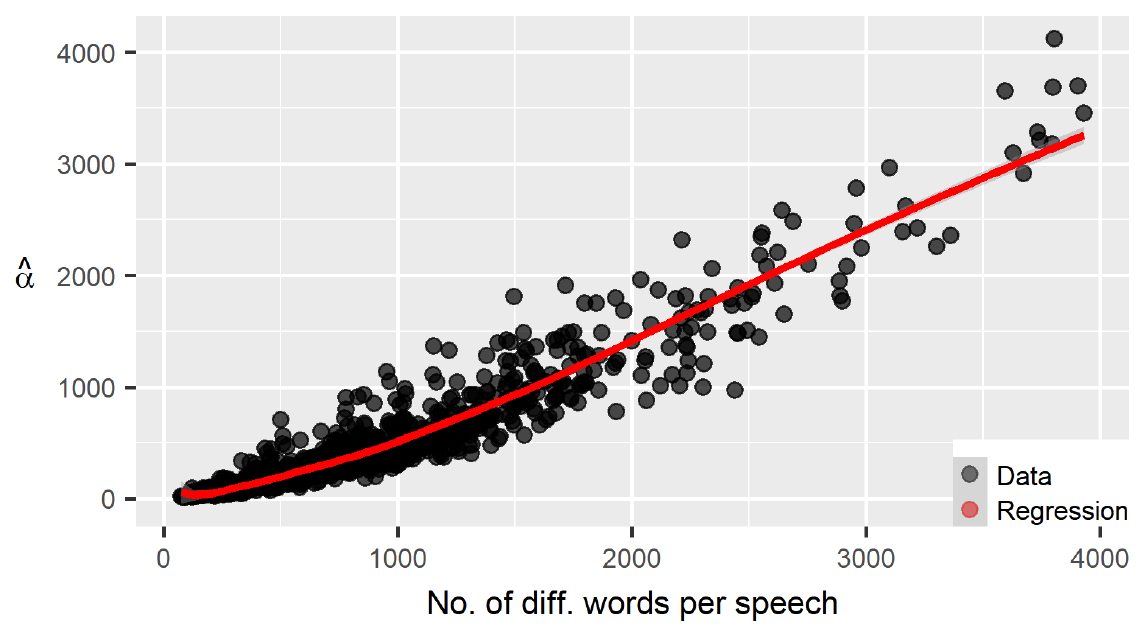}
\caption{Relationship between
$\hat{\alpha}$ and number of different words used in each speech (each word considered just once per corpus)}
\label{Fig6}
\end{figure}

\begin{figure}[h]
\centering
\includegraphics[width=5.7in]{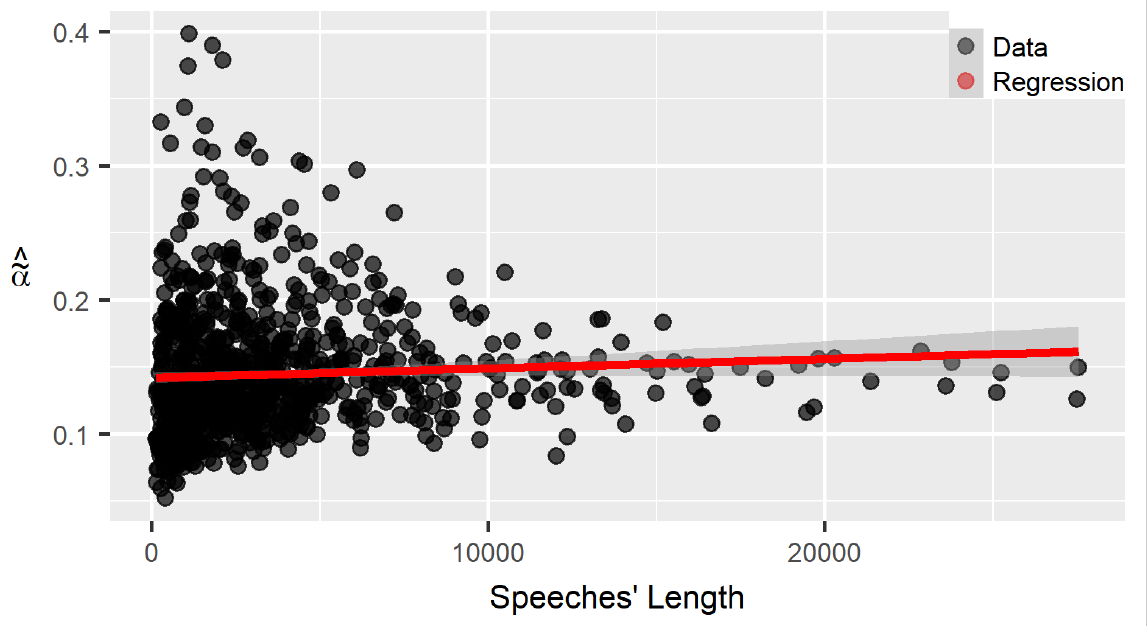}
\caption{Relationship between
$\hat{\tilde{\alpha}}$ calibrated for relative frequencies and speeches' length in term of total number of words used.}
\label{Fig7}
\end{figure}

\begin{figure}[h]
\centering
\includegraphics[width=5.7in]{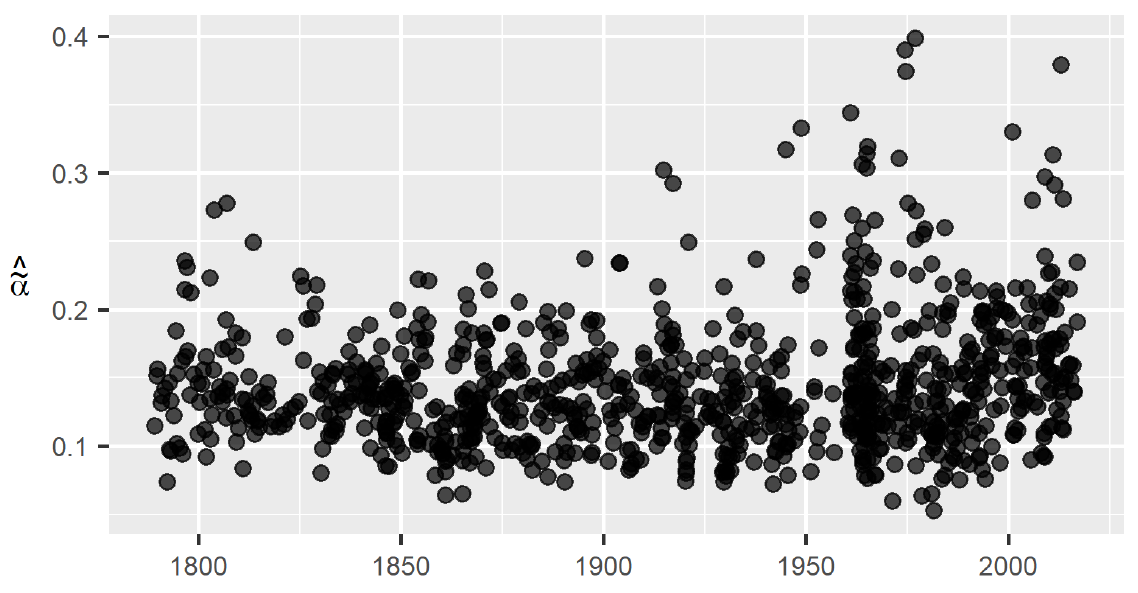}
\caption{Estimated $\hat{\tilde{\alpha}}$ on relative frequencies for each
speech over years (Eq. (\ref{ZML_rel})).}
\label{Fig8}
\end{figure}

\begin{figure}[h]
\centering
\includegraphics[width=5.7in]{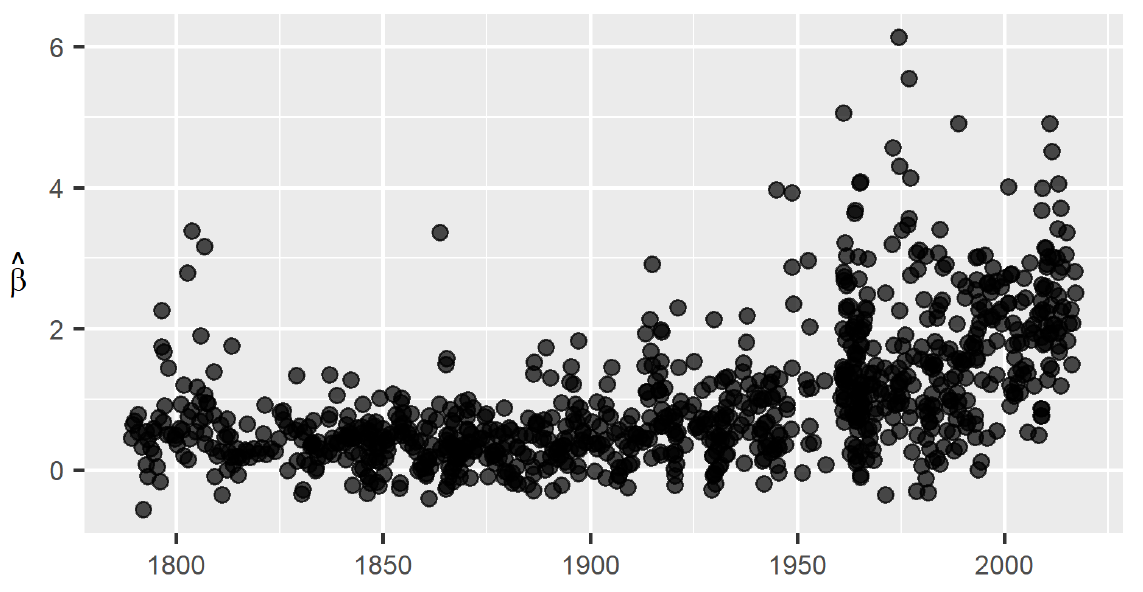}
\caption{Estimated $\hat{\beta}$ on relative frequencies for each
speech over years (Eq. (\ref{ZML_rel})).}
\label{Fig9}
\end{figure}

\begin{figure}[h]
\centering
\includegraphics[width=5.7in]{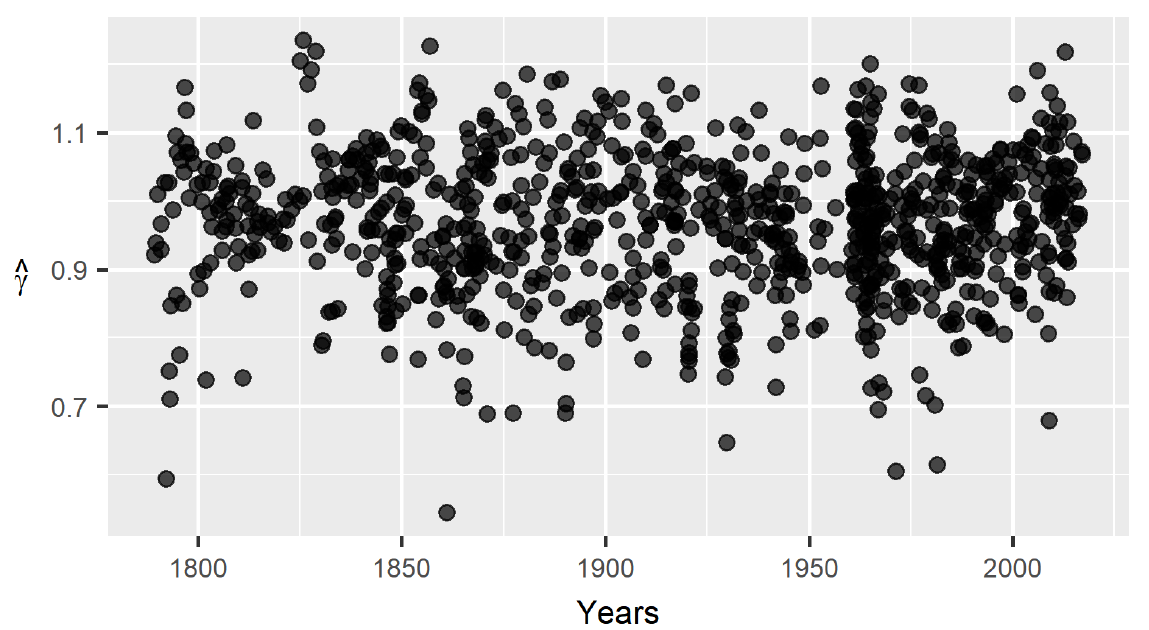}
\caption{Estimated $\hat{\gamma}$ on relative frequencies for each
speech over years (Eq. (\ref{ZML_rel})).}
\label{Fig10}
\end{figure}

\begin{figure}[h]
\centering
\includegraphics[width=5.7in]{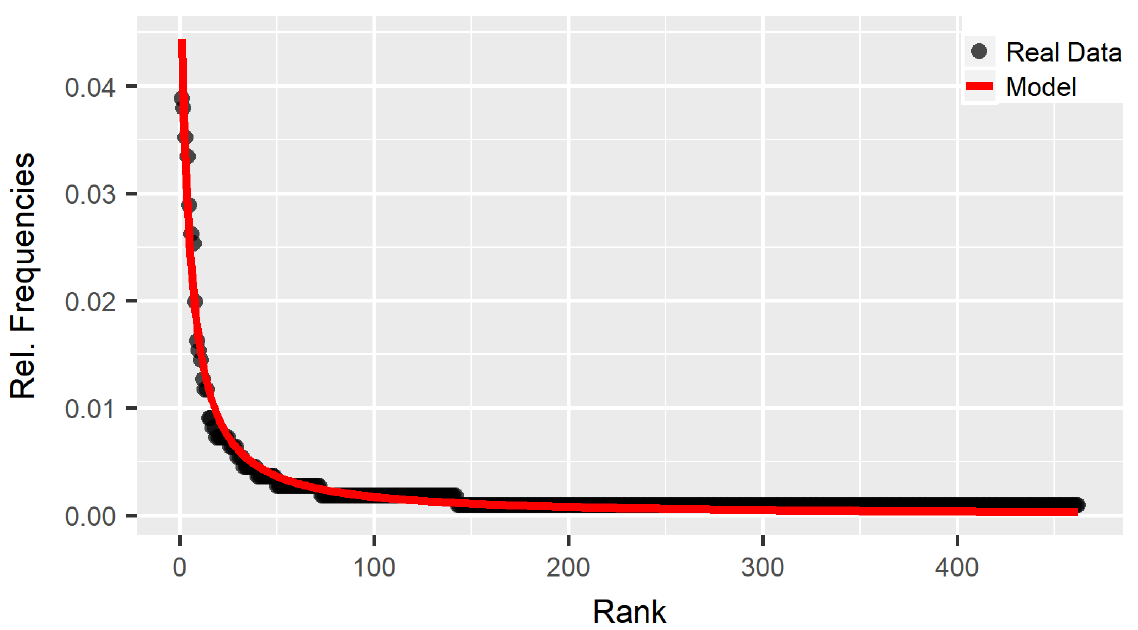}
\caption{January 20, 1977 - Inaugural Address - Jimmy Carter. Comparisons between real data and fitted models over the
speech's words relative frequencies for the case of the highest
$\hat{\tilde{\alpha}}= 0.39$; $\hat{\beta} = 5.54$; $\hat{\gamma} =
1.16$; N = 1107; $R^2=0.97$.}
\label{Fig11}
\end{figure}

\begin{figure}[h]
\centering
\includegraphics[width=5.7in]{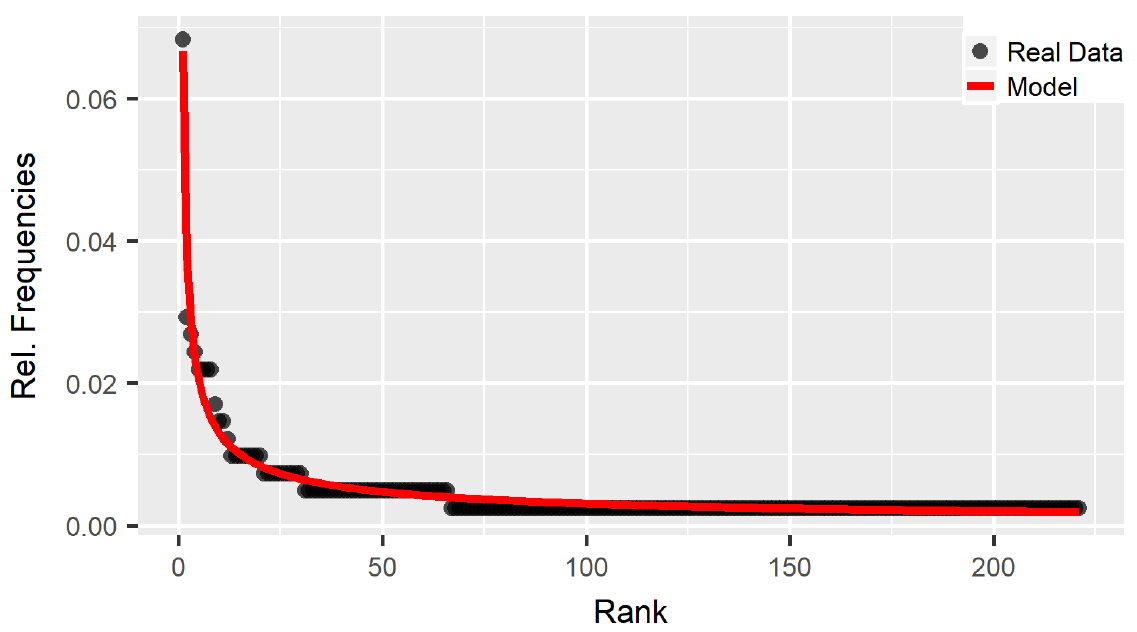}
\caption{August 3, 1981 - Remarks on the Air Traffic Controllers
Strike - Ronald Reagan. Comparisons between real data and fitted
models over the speech's words relative frequencies for the case of
the lowest $\hat{\tilde{\alpha}} = 0.05$; $\hat{\beta} = -0.32$;
$\hat{\gamma} = 0.61$; N = 410 $R^2=0.97$.} \label{Fig12}
\end{figure}

A visual inspection of $\hat{\alpha}$ in Figure \ref{Fig2} shows a
remarkable trend in the first years. Figure \ref{Fig3} evidences the
positive correlation between $\hat{\alpha}$ and the speech length
$N$ over all the speeches, while Figure \ref{Fig6} shows that the
correlation between $\hat{\alpha}$ and number of different tokens
used in each talk is still present, but is less clear than the
previous one. The dependence of $\hat{\alpha}$ on $N$ represents a
bias for the analysis of the results and a supportive argument for
studying also relative frequencies. In fact, such a dependence
disappears in calibrating the parameters with Eq. (\ref{ZML_rel}),
i.e. by taking into consideration the relative frequencies of the
words, as shown in Figure \ref{Fig7}. The calibrated parameters on
relative frequencies are reported in Figures
\ref{Fig8},\ref{Fig9},\ref{Fig10}. A statistical summary of the so
obtained parameter values is reported in Table
\ref{summarystatisticspar}; the goodness of fit measures is reported
in Table \ref{summarystatisticrsq}. The salient cases of
$\hat{\tilde{\alpha}}$ are presented in Figure \ref{Fig11},
\ref{Fig12}. The last one is associated to Ronald Reagan's speech
titled "Remarks on the Air Traffic Controllers Strike", which is
very short, thus having such a small $\hat{\tilde{\alpha}}$. We need
to say that the original transcript was longer; we remove them as
discussed in the second section. Thus, the parameter
$\tilde{\alpha}$ can be viewed still as an indicator of the highest
relative frequency in  each speech, even if its magnitude is
mitigated by the relationships with the other parameters.

\begin{figure}[h]
\centering
\includegraphics[width=5.7in]{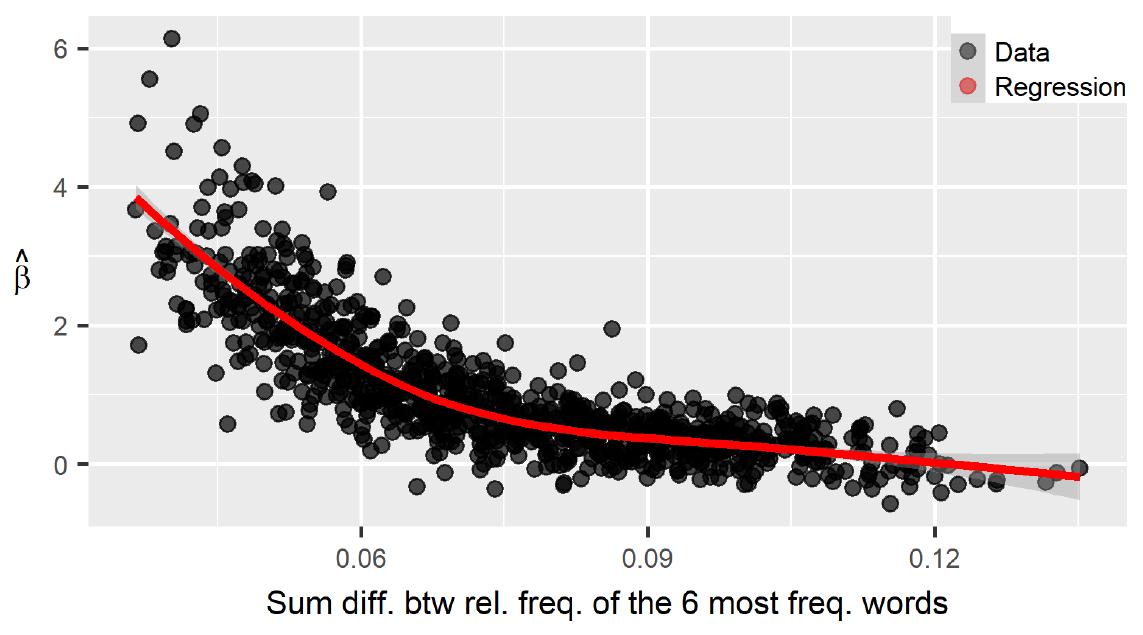}
\caption{$\hat{\beta}$ against the summed differences in relative frequencies of the top 6 repeated words within each speech.}
\label{Fig13}
\end{figure}

\begin{figure}[h]
\centering
\includegraphics[width=5.7in]{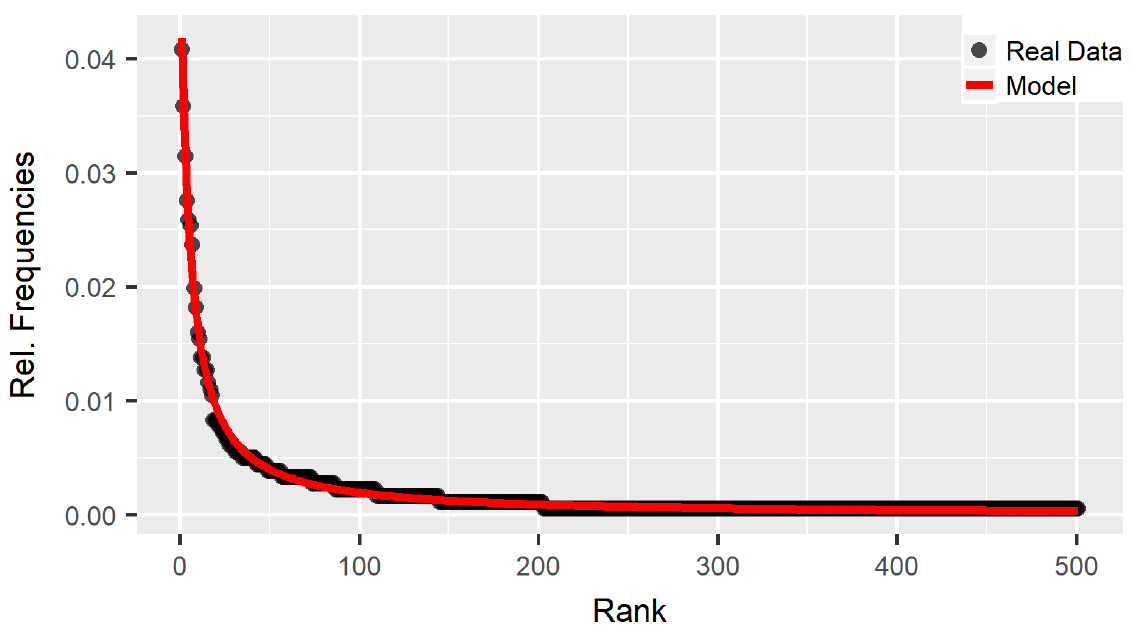}
\caption{August 9, 1974 - Remarks on Departure From the White House - Richard Nixon. Comparisons between real data and fitted models over the
speech's words relative frequencies for the case of the  highest $\hat{\beta} = 6.12$; $\hat{\tilde{\alpha}} =
0.38$; $\hat{\gamma} =1.13$; N = 1815; $R^2=0.99$.}
\label{Fig14}
\end{figure}

\begin{figure}[h]
\centering
\includegraphics[width=5.7in]{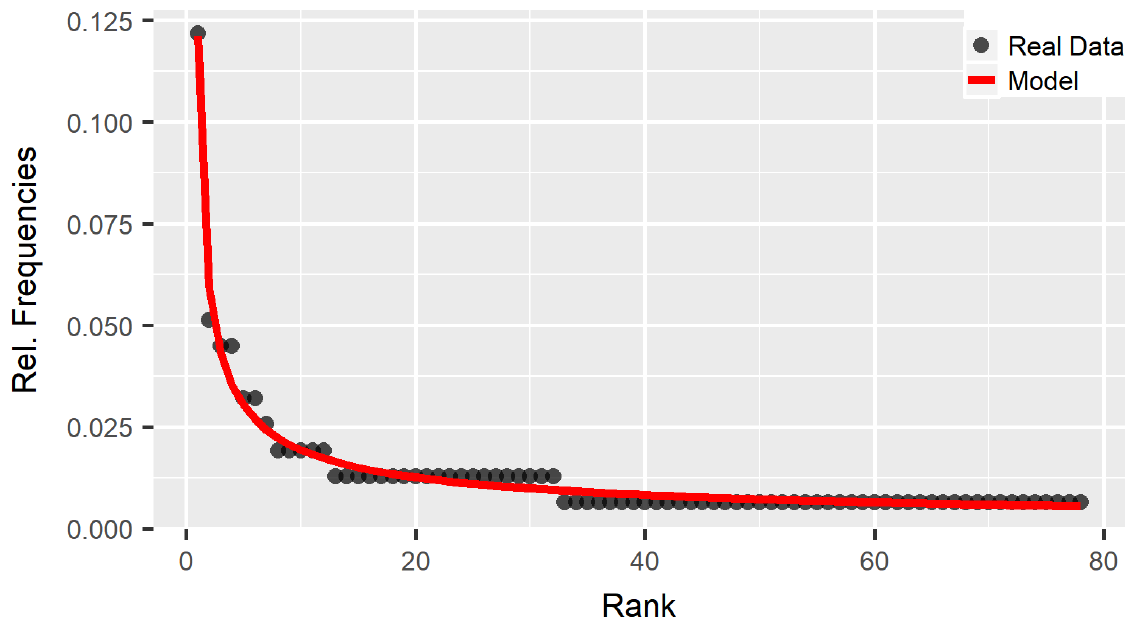}
\caption{April 5, 1792 - Veto Message on Congressional Redistricting -  George Washington. Comparisons between real data and fitted models over the
speech's words relative frequencies for the case of the lowest
$\hat{\beta} = -0.56$; $\hat{\tilde{\alpha}} = 0.07$; $\hat{\gamma}
= 0.59$; N = 156; $R^2=0.98$.}
\label{Fig15}
\end{figure}

The calibrated $\hat{\beta}$ gives  an  indication on the differences
among the frequencies within the various  speeches. Given that, the biggest
differences are between the low ranked words as well know by
\cite{zipf1949human,zipf1935psycho}. The $\hat{\beta}$s are showing
the behaviour of the frequencies at the lowest ranks, thus on the
most common words.
An evidence of the feature of $\hat{\beta}$ is provided from establishing  the differences
between words' frequencies at consecutive ranks within each speech.
Indeed, by summing for each speech the first 5 differences
originated by the 6 most repeated words and comparing them to the
$\hat{\beta}$s, one has Figure \ref{Fig13}. Such a
Figure shows the decay of $\hat{\beta}$ with respect to the
differences in frequencies within each speech. The graph is
confirming that high level of $\hat{\beta}$ is corresponding to tiny
differences between top six repeated words within speeches (see also
Figure \ref{Fig14}) while the converse occurs for the low
level of the parameters (see Figure \ref{Fig15}). This
occurrence can be interpreted also in terms of the rhetoric
structure, by asserting that when $\hat{\beta}$ is large, the words
of the speech have a more homogeneous distribution along the highest
ranks, hence pointing to the presence of a "rich club" of outliers at the low ranks.

\begin{figure}[h]
\centering
\includegraphics[width=5.7in]{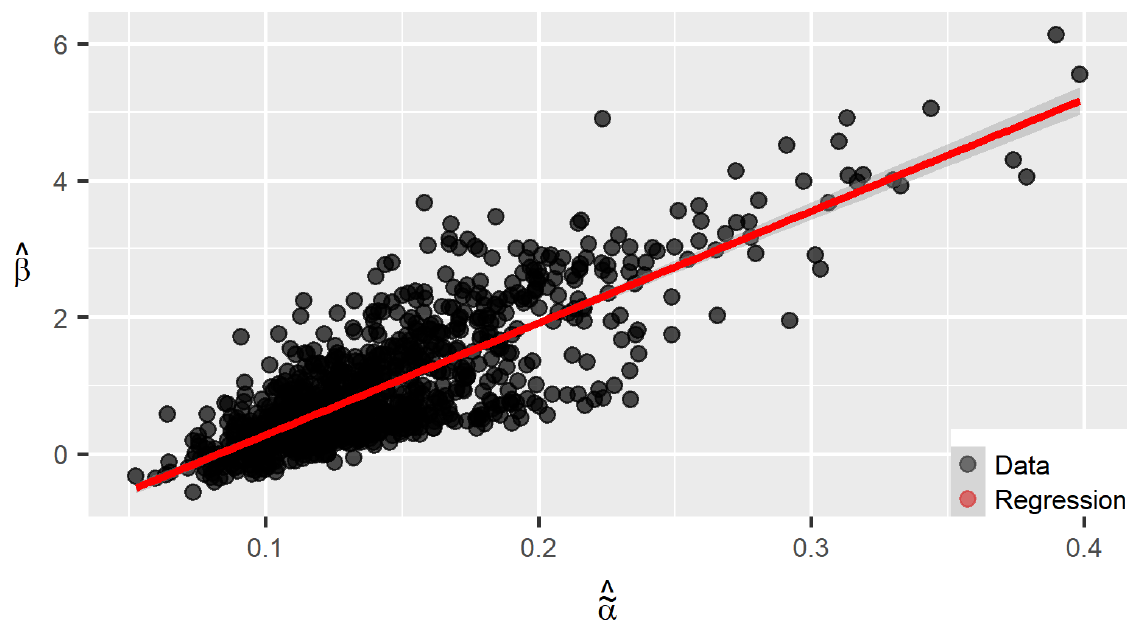}
\caption{Graphical insight of the relationship between $\hat{\tilde{\alpha}}$ and $\hat{\beta}$ in the estimation run using Eq. (\ref{ZML_rel}).}
\label{Fig16}
\end{figure}

Another interesting feature that emerges by the best fit procedure
on Eq. (\ref{ZML_rel}) is the correlation between
$\hat{\tilde{\alpha}}$ and $\hat{\beta}$ (see Figure
\ref{Fig16}). The joint evaluation of such calibrated
parameters gives then information on the magnitude of frequencies at
low ranks.

\begin{figure}[h]
\centering
\includegraphics[width=5.7in]{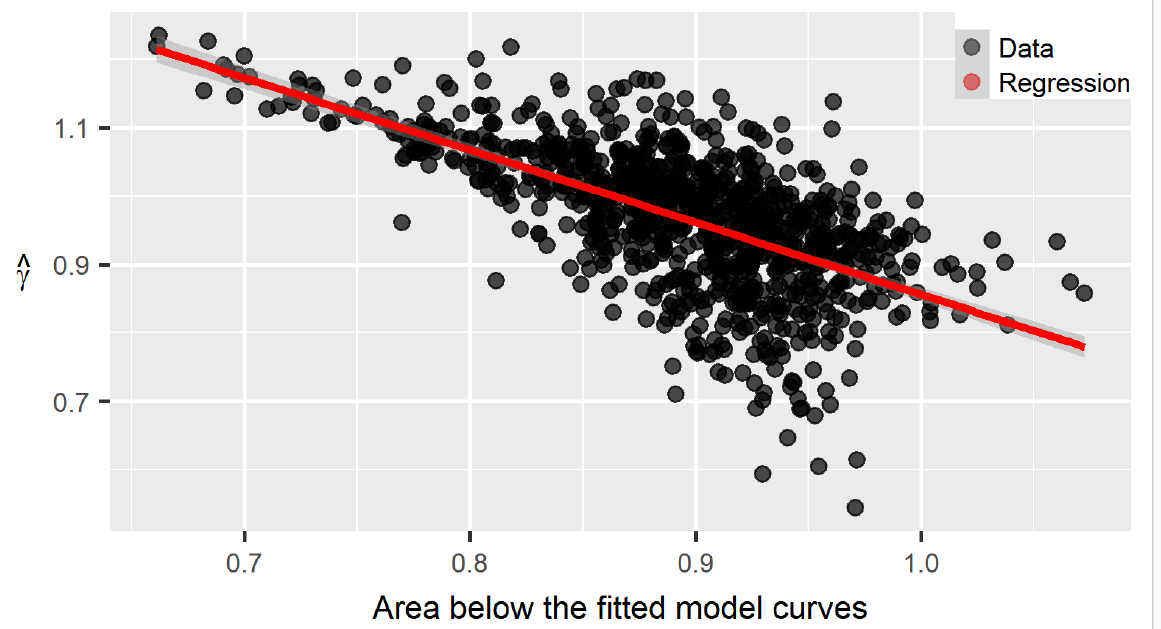}
\caption{The $\hat{\gamma}$ against the areas underlined by each fitted model computed with Eq. (\ref{integral}).}
\label{Fig17}
\end{figure}

\begin{figure}[h]
\centering
\includegraphics[width=5.7in]{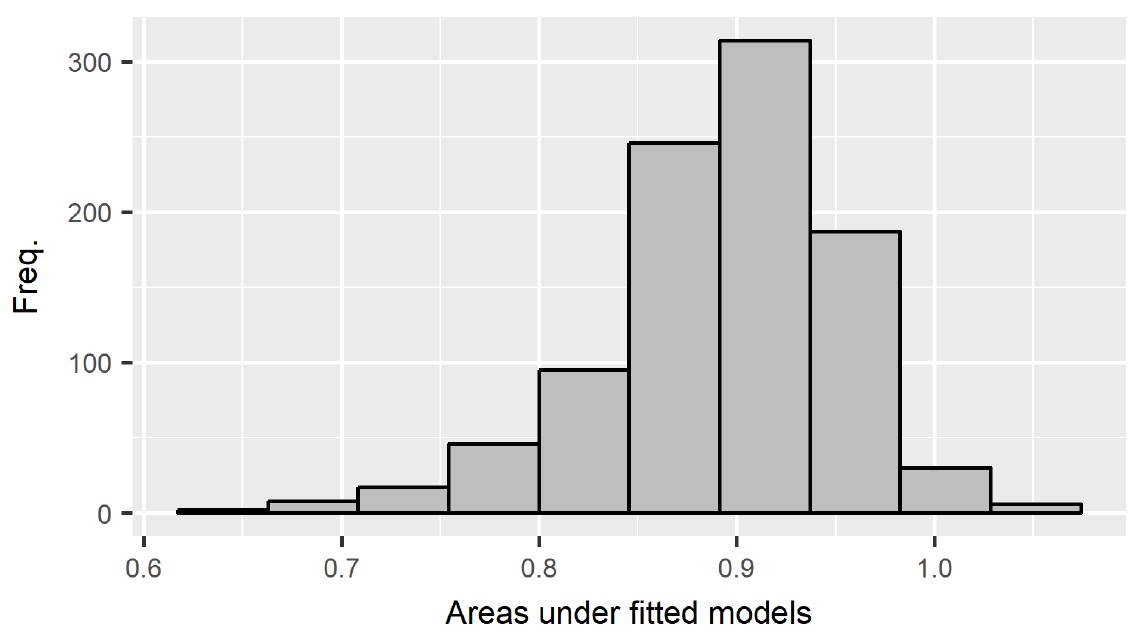}
\caption{Areas under the fitted models computed with Eq. (\ref{integral}).}
\label{Fig18}
\end{figure}

The concavity of the fitted curve related to Eq. (\ref{ZML_rel}) is mainly affected by $\hat{\gamma}$. Therefore this parameter is
informative about the decay of the model passing by the low ranks to
the high ranks. The $\hat{\gamma}$ is peculiar of each speech,
changing the focus of the hyperbolas in agreement with the features of
the talks. Consequently, it is the parameter that mostly affects the
areas under the fitted models, which is reduced when $\hat{\gamma}$
increases (see Figure \ref{Fig17} for a graphical
representation of the relationship between areas and
$\hat{\gamma}$). The areas have been calculated by computing the
following integral over each model characterized by
$\hat{\tilde{\alpha}}$, $\hat{\beta}$ and $\hat{\gamma}$, where the
$i$ is the indicator of the $i^{th}$ speech, so that $i=1, \ldots,
951$:
\begin{equation}
\label{integral}
    A_i = \int_{1}^{r_{Max,i}}\hat{\tilde{\alpha}}_i(r+\hat{\beta}_i)^{-\hat{\gamma}_i} dr = \frac{\hat{\tilde{\alpha}}_i}{1-\hat{\gamma}_i}\left[(r_{Max,i}+\hat{\beta}_i)^{1-\hat{\gamma}_i}-(1+\hat{\beta}_{i})^{1-\hat{\gamma}_i}\right]
\end{equation}
$A_i$ is the $i^{th}$ area corresponding  to the model calibrated
over the $i^{th}$ talks, while   $r_{Max,i}$ is the highest rank
referred to the $i^{th}$ transcript. Refer to Figure \ref{Fig18},
where it is possible to notice the shape of the histograms defined
in Eq. (\ref{ZML_rel}). Such a Figure gives a clear idea of the ZML
capacity of being a density function in this specific case of the
analysis of relative frequencies.

\begin{figure}[h]
\centering
\includegraphics[width=5.7in]{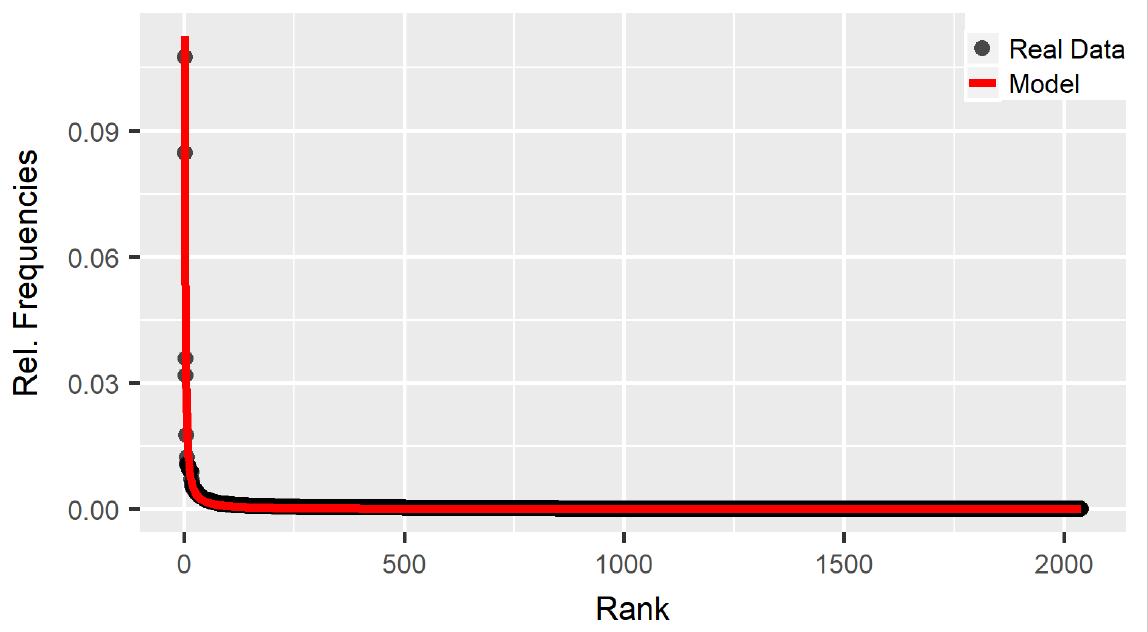}
\caption{December 6, 1825 - First Annual Message - John Quincy Adams. Comparisons between real data and fitted models over the
speech's words relative frequencies for the case of the highest $\hat{\gamma} = 1.23$;
$\hat{\tilde{\alpha}}= 0.21$; $\hat{\beta} = 0.70$; N = 9023;
$R^2=0.96$.}
\label{Fig19}
\end{figure}

\begin{figure}[h]
\centering
\includegraphics[width=5.7in]{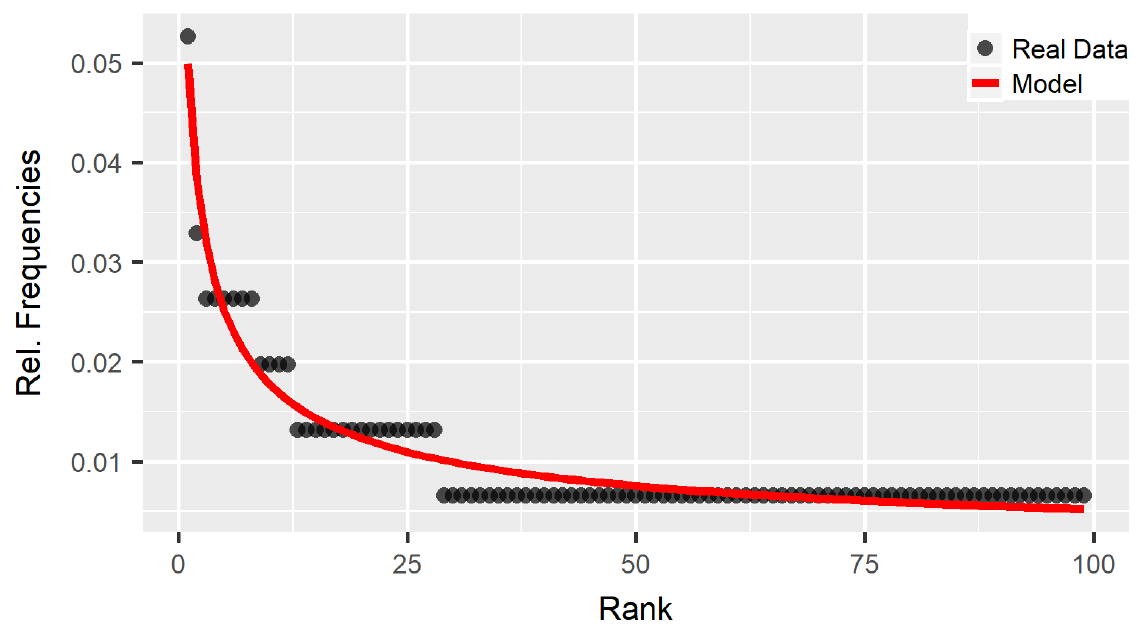}
\caption{"February 11, 1861 - Farewell Address - Abraham Lincoln. Comparisons between real data and fitted models over the
speech's words relative frequencies for the case of the  lowest $\hat{\gamma} = 0.54$; $\hat{\tilde{\alpha}}= 0.06$; $\hat{\beta} = 0.58$; N = 152; $R^2=0.93$.}
\label{Fig20}
\end{figure}

Notice also that the $\hat{\gamma}$ is low (high) when in the speech
there is a more or less evident transition  from low ranked words to
high ranked words; see the two cases in Figures
\ref{Fig19}, \ref{Fig20}.

\begin{figure}[h]
\centering
\includegraphics[width=5.7in]{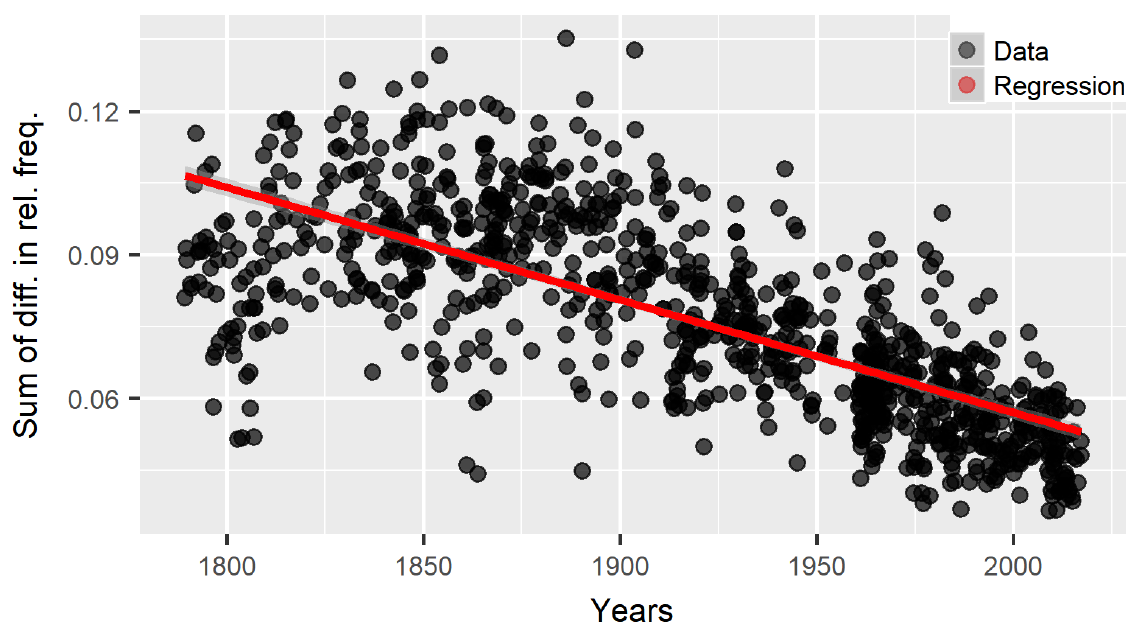}
\caption{Sum of the differences computed between
all the words' relative frequencies within each speech along the
years.}
\label{Fig21}
\end{figure}

Under an evolutive perspective, the most pronounced words for each
speech have a constant decreasing rate along the years (see Figure
\ref{Fig23}). This result means that the repetition
of a single word
tends to be reduced with time. This occurrence goes hand in hand
with a global reduction in the differences between the words'
frequencies within the speeches, as shown  in Figure \ref{Fig21}.
Considering that the most pronounced tokens are mainly conjunctions,
articles and preposition, this phenomenon could be interpreted as
the growing need in time of making simpler syntax of the sentences.
Another useful hint of this fact is that the number of single words
used (see Figure \ref{Fig1}) is slightly decreasing along the years.
So the global tendency is to use less single words and propose
sentence structures which are not too complex.

\begin{table}[h]
\centering
\begin{tabular}{rrrrrrr}
  \hline
 & $\hat{\alpha}$ & $\hat{\beta}$ & $\hat{\gamma}$ & $\hat{\tilde{\alpha}}$ & $\hat{\beta}$ & $\hat{\gamma}$  \\
  \hline
  Max & \numprint{4117.92} & 6.13 & 1.23 & 0.40 & 6.13 & 1.23  \\
  Min & 9.72 & -0.57 & 0.54 &  0.05 & -0.57 & 0.54 \\
  Median $m$ & 326.35 & 0.72 & 0.9 & 0.14 & 0.72 & 0.97\\
  Mean $\mu$ & 521.13 & 1.01 & 0.97 & 0.14 & 1.01 & 0.97 \\
  RMS & 25.37 & 0.05 & 0.03 & 0.004 & 0.05 & 0.03\\
  Standard Deviation & 583.69 & 0.97 & 0.10 & 0.05 & 0.97 & 0.10\\
  Variance & \numprint{340339.10} & 0.94 & 0.01 & 0.002 & 0.94 & 0.01\\
  Standard Error & 18.93 & 0.03 & 0.003 & 0.001 & 0.03 & 0.003\\
  Skewness  & 2.40 & 1.41 & -0.45 & 1.55 & 1.41 & -0.41 \\
  Kurtosis & 10.40 & 5.35 & 3.65 &  6.90 & 5.36 & 3.53 \\
  \hline
  $\mu/\sigma$ & 0.89 & 1.03 & 9.53 & 3.04 & 1.04 & 9.53\\
  $3(\mu-m)/\sigma$ & -1.00 & -0.89 & 0.20 & -0.56 & -0.89 & 0.20\\
   \hline
\end{tabular}\caption{Statistical summary of the estimated parameters on relative
and absolute frequencies in accordance with Eqs. (\ref{ZML}) and
(\ref{ZML_rel}) respectively. } \label{summarystatisticspar}
\end{table}

\begin{table}[h]
\centering
\begin{tabular}{rcccc}
  \hline
 & $R^2$ & Std error & $R^2_{rel}$ & Std error$_{rel}$\\
  \hline
Max & 1.00 & 5.68 & 1.00 & 0.0041\\
  Min & 0.91 & 0.26 & 0.91 & 0.0001\\
  Median $m$  & 0.99 & 1.00 & 0.99 & 0.0004\\
  Mean $\mu$ & 0.98 & 1.25 &0.98 & 0.0006 \\
  Standard Deviation & 0.01 & 0.87 & 0.01 & 0.0005 \\
   \hline
\end{tabular}
\caption{Statistical summary of $R^2$s and non-linear regression
standard errors calculated for each fit with Eqs. (\ref{ZML}) and
(\ref{ZML_rel}). They represent the models goodness of fit
calibrated over each speech when considering absolute and relative
frequencies.} \label{summarystatisticrsq}
\end{table}

\begin{table}[ht]
\centering
\begin{tabular}{rcc}
  \hline
 & S.W. p-values of fits on Eq. (\ref{ZML})& S.W. p-values of fits on Eq. (\ref{ZML_rel})\\
  \hline
 Max &0.04 & 0.04\\
 Min & 0.00   & 0.00\\
 Median $m$  & 0.00  & 0.00  \\
 Mean $\mu$  & 0.00   & 0.00\\
 Standard Deviation & 0.00 & 0.00 \\
 \hline
\end{tabular}
\caption{Statistical summary of the Shapiro--Wilk test (S.W. in the
Table) p-values resulting from the tests performed on standardized
residuals of best fit run with Eqs. (\ref{ZML}) and
(\ref{ZML_rel}).} \label{summarypvalues}
\end{table}


\section{Discussion and conclusive remarks}
\label{Conclusion} The Presidential speeches' sample analyzed in
this paper is one of the most complete in the literature. It has
been constructed under consistency criteria in a phase-wise form;
finally it contains 951 talks over a span of about 228 years  for
all the US Presidents up to now. The source of the data is the
Miller Center website;  data have been retrieved at the end of June
2017.

The use of rank-size laws with sizes given by the word frequencies
through the Zip-Mandelbrot law (see Eq. (\ref{ZML}) and Eq.
(\ref{ZML_rel})) gives the opportunity of analyzing the rhetoric
structure of the transcripts. More specifically, the method allows
us to observe changes into  the rhetoric frameworks without being
drastically affected by the changes in usages and significance of
terminologies that occurred during the years. Indeed such an
objective analysis of the frequencies disregards the words' meanings
and focuses on the exploration of the structures of the speeches
when words are ranked. Undoubtedly, over the 228 years hereby
considered, the Americans' language has changed a lot as well as the
"society" at which the speeches are addressed; thus,   the
convergence toward a common scheme in producing talks could be even
explored searching for "universal behavior". Thus, Eqs. (\ref{ZML})
- (\ref{ZML_rel}) show a spectacular capacity of fitting the
transcripts, but the considerations on regressions standard errors
and values of $R^2$ always around 1 (see Table
\ref{summarystatisticrsq} and Figure \ref{Fig5}) lead to the
conclusion that Eq. (\ref{ZML_rel}) is more suitable to represent
the talks, especially considering the bias on $\hat{\alpha}$.
Further evidence of the impressive  capacity of such a fitting can
be noticed with a visual inspection of Figures
\ref{Fig11},\ref{Fig12},\ref{Fig14},\ref{Fig15},\ref{Fig19},\ref{Fig20}.
Therefore, jointly with the Shapiro-Wilk test p-values reported in
Table \ref{summarypvalues}, one can conclude  that the selected
approach does not reveal compromising weaknesses, can be considered
rigorous, and recommended in further works.

The calibrated parameters on Eq. (\ref{ZML_rel}) are presented in
Figures \ref{Fig8},\ref{Fig9},\ref{Fig10}. It is possible to observe some
changes along the years by visual inspecting $\hat{\tilde{\alpha}}$,
$\hat{\beta}$ and $\hat{\gamma}$. Such calibrated parameters are
used to resume some features of the speeches structures. 

The $\hat{\tilde{\alpha}}$ of Eq. (\ref{ZML_rel}) has a small
increment in volatility during the last years with a high
concentration of outliers after the 1960s. Considering the fact that
$\tilde{\alpha}$ is giving an  indication on the  relative
frequencies of the most often used words, the meaning of the related
behavior along the years can be interpreted as an upcoming of
irregularities in the use of words.

The analysis of $\hat{\alpha}$, so the parameter estimated with Eq.
(\ref{ZML}), whose behavior is reported in Figure \ref{Fig2},
leads to the assessment of two remarkable trends of speeches' length during the years between 1800-1850 and 1850-1900. As we have said before,
this outcome is grounded on the fact that the
parameter $\alpha$ can be considered    as  a proxy for exploring the
number of words employed in the speech.

The $\hat{\beta}$ has a similar behavior to that  of
$\hat{\tilde{\alpha}}$, as   can be seen from Figure
\ref{Fig16}. Indeed, its points are quite homogeneously
distributed between 0 and 1 until 1900 when $\hat{\beta}$
starts to rise with a contemporaneous increment of  the volatility (see Figure \ref{Fig9}).

\begin{figure}[h]
\centering
\includegraphics[width=5.7in]{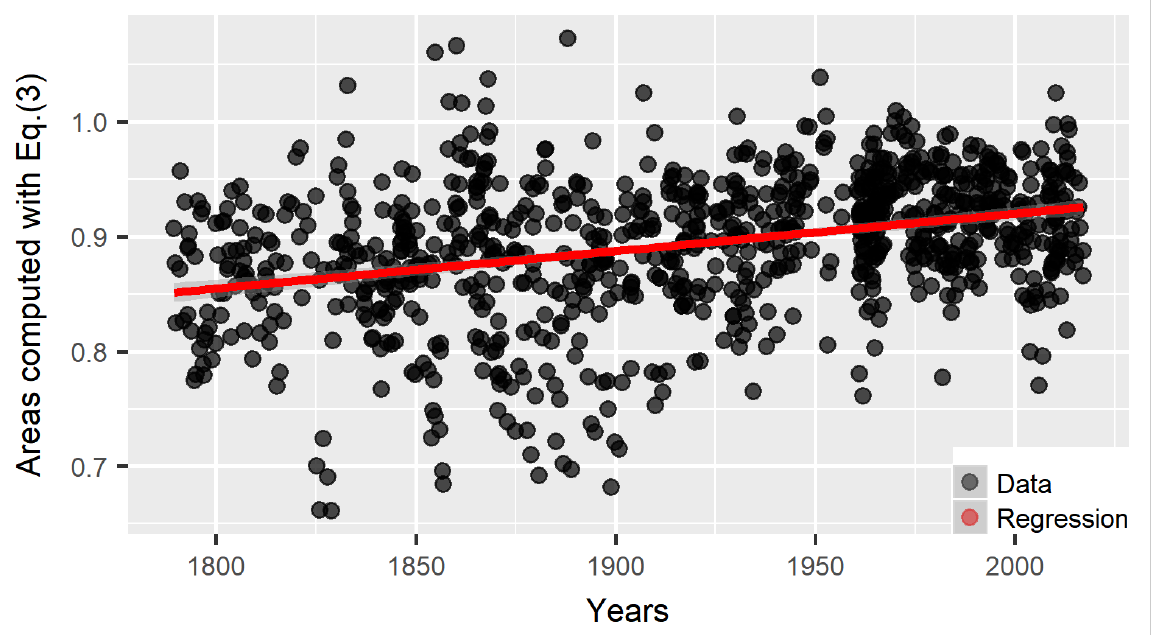}
\caption{Each point represents an area under the respective fitted
model computed with Eq. (\ref{integral}).}
\label{Fig22}
\end{figure}

\begin{figure}[h]
\centering
\includegraphics[width=5.7in]{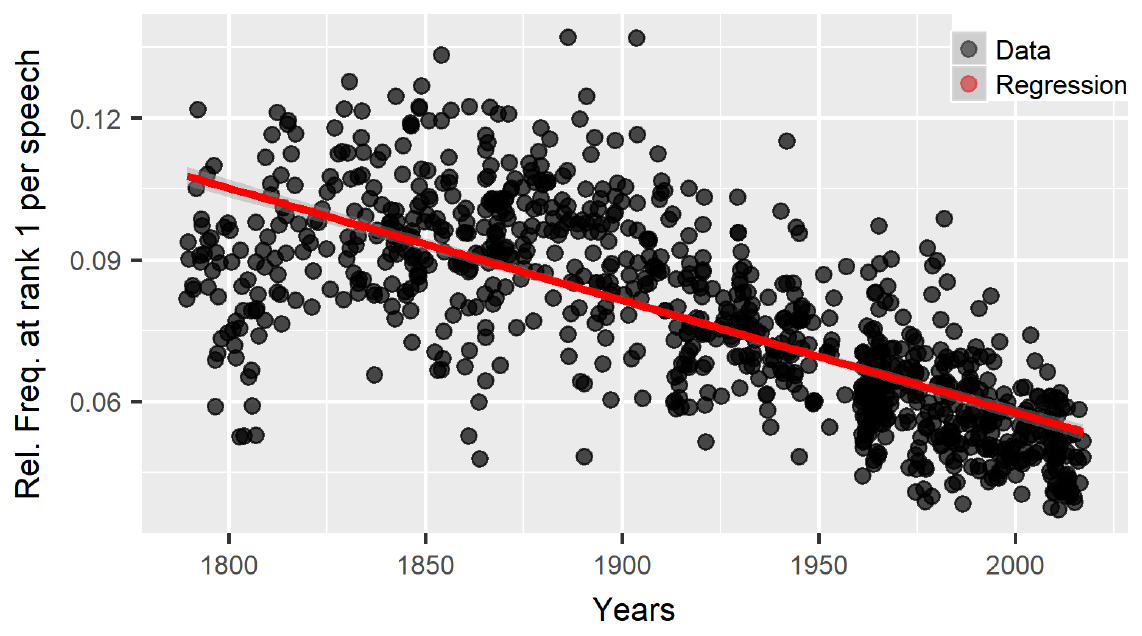}
\caption{The relative frequencies of the most used word in each
speech along the years.}
\label{Fig23}
\end{figure}

The $\beta$'s increment when the differences between frequencies
at low ranks are decreasing (see Figure \ref{Fig13}) helps
us to conclude that after 1900 the words  frequencies distributions
are converging toward more homogeneous distributions. A further
confirmation of this is given by the areas delimited by the models
and computed through Eq. (\ref{integral}). As it is possible to
deduce from Figure \ref{Fig22}, there is a feeble positive
trend, combined with a reduction in variability. Furthermore, there
is a clear decreasing trend in the most often used words' relative
frequencies of each speech (see Figure \ref{Fig23}), which
reinforces the results and interpretations  about  $\hat{\beta}$.

\begin{figure}[h]
\centering
\includegraphics[width=5.7in]{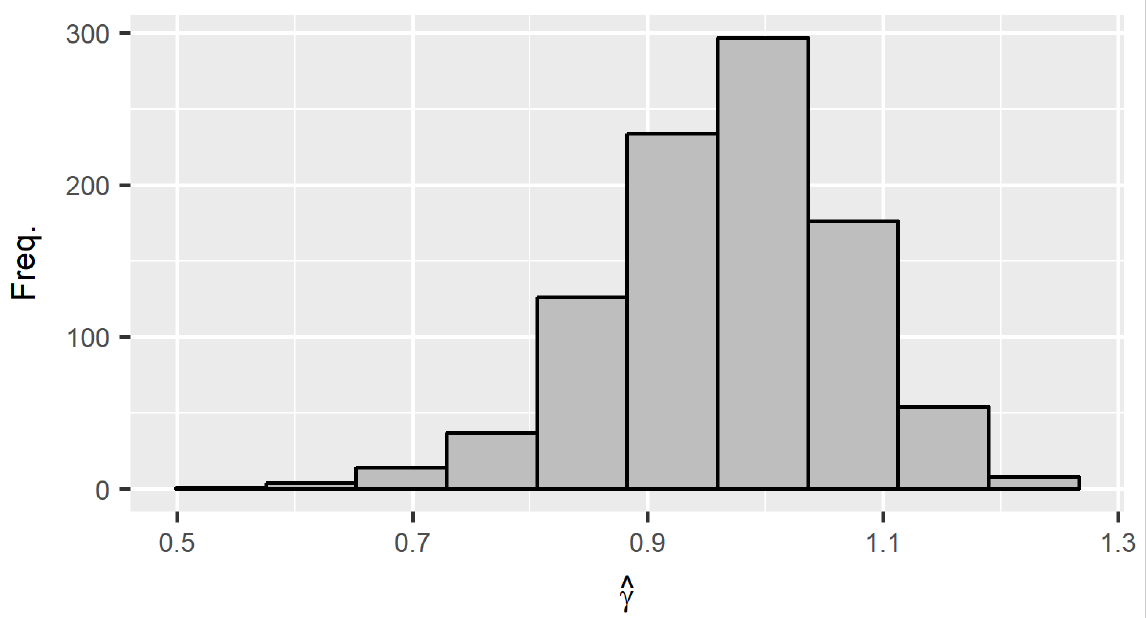}
\caption{Histogram of the $\hat{\gamma}$' values.}
\label{Fig24}
\end{figure}

From Figure \ref{Fig10}, the calibrated $\hat{\gamma}$ appears to be
quite stable in terms of trend and distance from 1. In the majority
of the cases, such a parameter assumes a value around 1. When
$\hat{\gamma}\ge 1$, then one can assert that there is
a steeper decay of the data in the rank-size plot. Figure
\ref{Fig24} assists in visualizing that the distribution of
the $\hat{\gamma}$'s is asymmetric and the left tail is a bit longer than the right tail, 
giving a further indication of the tendency toward President producing "more homogeneous
talks", in our sense.

Finally we can assert that the speeches' structures exhibit in
general a sort of common framework, with a specific proportion of
words. Consequently, this means that the typology of the rhetoric
involved in the political public speaking is identifiable. This can
be considered as a supporting argument of the generic structural
system that lies behind the rhetoric of American Presidents'
political speeches. Moreover, patterns in the parameters behaviors
can be viewed as an hint about a common generative womb of the
speeches rhetoric. The method here applied is informative and the
robust capacity of fitting provides a certain confidence in reaching
a conclusion.

The obtained results suggest the presence of common linguistics
thoughts of the Presidents when delivering a talk. One can argue
that Presidents imitate their predecessors and look at the speeches
already delivered to get inspiration for their own talks. The
historical discrepancies point to the presence of periodical modes
of the structure of the talks and on the linguistic productivity.
Such an outcome should be correctly interpreted not only in terms of
individual Presidents, but also accordingly to the evolution of the
language. The regularity over all the Presidents of the decay of the
best-fit curve from the high ranks to the low ones can be viewed as
an evidence of the similarities among the speeches. Such an outcome
suggests that the Presidents' speeches share the same way to use
words and their frequencies, hence pointing to a sort of coded
communication channel when delivering talks.

All these elements impose rigorous and deep investigations, and are
opening the door for further studies.

One of the most prominent proposal for future research concerns the
assessment of the stochastic properties of the rhetoric of the
speeches for performing some forecast of the structure of future
President speeches. A further research theme is associated to the
way Presidents' speeches might be clustered. In this respect, it is
possible to propose different concepts of distance between couples
of speeches bringing different typologies of information, and
identify accordingly the closest speeches. Results might be
interpreted at the level of individual Presidents or in the specific
historical context. This proposal suggests to analyze also subsets
of speeches delivered in specific situations (like war periods or
times of economic distress) or by specific Presidents to identify
sources of regularities and guess the presence of imitative
behaviors or coded communication rules.

Under a pure methodological perspective, the proposed combination of
text-mining and rank-size approaches exhibits characteristics of
flexibility and generality, hence suggesting the possibility of an
effective reproduction of the analysis in other relevant expert
systems contexts. In this respect, the analysis of the speeches of
the Governors of the Central Banks might be quite informative on the
way in which the different phases of business cycles and remarkable
economic situations are treated by leading economists with a so
relevant institutional role.

\nolinenumbers
\clearpage

\bibliographystyle{apalike}

\clearpage
\begin{thebibliography}{99}
\bibitem{alduy} Alduy, C. (2017). Ce qu’ils disent vraiment. Les politiques pris aux mots. Le Seuil.



\bibitem{vejdemo2016semantic}  Vejdemo, S. and Horberg, T. (2016). Semantic factors predict the rate of lexical replacement of content
words. PloS one, 11(1):e0147924.
\bibitem{tetlock2008more}  Tetlock, P. C., Saar-Tsechansky, M., and Macskassy, S. (2008). More than words: Quantifying language to
measure firms' fundamentals. The Journal of Finance, 63(3):1437-1467.
\bibitem{amrit}  Amrit, C., Paauw, T., Aly, R., and Lavric, M. (2017). Identifying child abuse through text mining and machine learning. Expert systems with applications, 88:402-418.
\bibitem{noh} Noh, H., Jo, Y., and Lee, S. (2015). Keyword selection and processing strategy for applying text mining to patent analysis. Expert Systems with Applications, 42(9):4348-4360.
\bibitem{nassirtoussi2014text} Nassirtoussi, A. K., Aghabozorgi, S., Wah, T. Y., and Ngo, D. C. L. (2014). Text mining for market prediction: A systematic review. Expert Systems with Applications, 41(16):7653-7670.
\bibitem{mostafa}  Mostafa, M. M. (2013). More than words: Social networks' text mining for consumer brand sentiments. Expert Systems with Applications , 40(10):4241-4251.
\bibitem{oberreuter2013text}  Oberreuter, G. and Velasquez, J. D. (2013). Text mining applied to plagiarism detection: The use of words
for detecting deviations in the writing style. Expert Systems with Applications , 40(9):3756-3763.
\bibitem{berezina} Berezina, K., Bilgihan, A., Cobanoglu, C., and Okumus, F. (2016). Understanding satisfied and dissatisfied hotel customers: text mining of online hotel reviews. Journal of Hospitality Marketing \& Management, 25(1):1-24.
\bibitem{moro}  Moro, S., Cortez, P., and Rita, P. (2015). Business intelligence in banking: A literature analysis from 2002 to 2013 using text mining and latent Dirichlet allocation. Expert Systems with Applications , 42(3):1314-1324.
\bibitem{pletsche}  Pletscher-Frankild, S., Pallejà, A., Tsafou, K., Binder, J. X., \& Jensen, L. J. (2015). DISEASES: Text mining and data integration of disease-gene associations. Methods, 74:83-89.
\bibitem{westergaard}  Westergaard, D., Straefeldt, H.-H., Tonsberg, C., Jensen, L. J., and Brunak, S. (2018). A comprehensive and quantitative comparison of text-mining in 15 million full-text articles versus their corresponding abstracts. 35 PLoS Computational Biology, 14(2):e1005962.
\bibitem{defortuny} De Fortuny, E. J., De Smedt, T., Martens, D., and Daelemans, W. (2012). Media coverage in times of political crisis: A text mining approach. Expert Systems with Applications, 39(14):11616-11622.
\bibitem{griffiths2004finding}  Grifiths, T. L. and Steyvers, M. (2004). Finding scientific topics. Proceedings of the National Academy of
Sciences , 101(suppl 1):5228-5235.
\bibitem{calude2011we} Calude, A. S. and Pagel, M. (2011). How do we use language? Shared patterns in the frequency of word use across 17 world languages. Philosophical Transactions of the Royal Society of London B: Biological Sciences, 366(1567):1101-1107.
\bibitem{ausloos2012generalized}  Ausloos, M. (2012a). Generalized Hurst exponent and multifractal function of original and translated texts
mapped into frequency and length time series. Physical Review E , 86(3):031108.
\bibitem{ausloos2012measuring}  Ausloos, M. (2012b). Measuring complexity with multifractals in texts. Translation effects. Chaos, Solitons \& Fractals , 45(11):1349-1357.
\bibitem{altmann2012origin}  Altmann, E. G., Cristadoro, G., and Degli Esposti, M. (2012). On the origin of long-range correlations in texts. Proceedings of the National Academy of Sciences , 109(29):11582-1158
\bibitem{zeng2012topics}  Zeng, J., Duan, J., Cao, W., and Wu, C. (2012). Topics modeling based on selective zipf distribution. Expert Systems with Applications , 39(7):6541-6546
\bibitem{milizia2014out}  Milizia, D. (2014). In, out, or half way? The European attitude in the speeches of British leaders. Lingue e Linguaggi , 11:157-175.
\bibitem{tetlock2007giving} Tetlock, P. C. (2007). Giving content to investor sentiment: The role of media in the stock market. The Journal of Finance, 62(3):1139-1168.
\bibitem{lim2002five}  Lim, E. T. (2002). Five trends in presidential rhetoric: An analysis of rhetoric from George Washington to Bill Clinton. Presidential Studies Quarterly , 32(2):328-348.
\bibitem{tang2013recognition} Tang, P. and Chow, T. W. (2013). Recognition of word collocation habits using frequency rank ratio and inter-term intimacy. Expert Systems with Applications, 40(11):4301-4314.
\bibitem{ATKINSON20134346}  Atkinson, J. and Munoz, R. (2013). Rhetorics-based multi-document summarization. Expert Systems with
Applications , 40(11):4346- 4352.
\bibitem{KRAUS201965} Kraus, M. and Feuerriegel, S. (2018). Sentiment analysis based on rhetorical structure theory:learning deep
neural networks from discourse trees. Expert Systems with Applications, 118:65-79.
\bibitem{saravanan2008automatic}   Saravanan, M., Ravindran, B., and Raman, S. (2008). Automatic identification of rhetorical roles using
conditional random fields for legal document summarization. In Proceedings of the Third International Joint Conference on Natural Language Processing: Volume-I .
\bibitem{SIDOROV2014853}  Sidorov, G., Velasquez, F., Stamatatos, E., Gelbukh, A., and Chanona-Hernandez, L. (2014). Syntactic n-grams as machine learning features for natural language processing. Expert Systems with Applications, 41(3):853- 860. Methods and Applications of Artificial and Computational Intelligence.

\bibitem{saravanan2006probabilistic} Saravanan, M., Raman, S., and Ravindran, B. (2006). A probabilistic approach to multi-document summarization for generating a tiled summary. International Journal of Computational Intelligence and Applications, 6(02):231-243.
\bibitem{stamatatos2009survey} Stamatatos, E. (2009). A survey of modern authorship attribution methods. Journal of the American Society
for information Science and Technology , 60(3):538-556.

\bibitem{ausloos2008equilibrium} Ausloos, M. (2008). Equilibrium and dynamic methods when comparing an English text and its Esperanto
translation. Physica A: Statistical Mechanics and its Applications, 387(25):6411-6420.
\bibitem{ausloos2010punctuation}  Ausloos, M. (2010). Punctuation effects in English and Esperanto texts. Physica A: Statistical Mechanics
and its Applications , 389(14):2835-2840
\bibitem{ausloos2016quantifying}  Ausloos, M., Nedic, O., Fronczak, A., and Fronczak, P. (2016). Quantifying the quality of peer reviewers
through Zipf's law. Scientometrics , 106(1):347-368
\bibitem{ferrer2010random}  Ferrer-i-Cancho, R. and Elvevag, B. (2010). Random texts do not exhibit the real Zipf's law-like rank
distribution. PLoS One , 5(3):e9411.
\bibitem{FEUERRIEGEL201888} Feuerriegel, S. and Gordon, J. (2018). Long-term stock index forecasting based on text mining of regulatory
disclosures. Decision Support Systems, 112:88-97.
\bibitem{herdan1958language} Herdan, G. (1958). Language as choice and chance. Philosophy and Phenomenological Research, 18 (4):565-565 
\bibitem{herdan1966advanced}  Herdan, G. (1966). The advanced theory of language as choice and chance, volume 4. Springer Berlin.
\bibitem{i2003least} Ferrer-i-Cancho, R. and Sole, R. V. (2003). Least effort and the origins of scaling in human language. Proceedings
of the National Academy of Sciences, 100(3):788-791.
\bibitem{ji2014representation}  Ji, Y. and Eisenstein, J. (2014). Representation learning for text-level discourse parsing. In Proceedings of the 52nd Annual Meeting of the Association for Computational Linguistics (Volume 1: Long Papers),  13-24.
\bibitem{piantadosi2014zipf}  Piantadosi, S. T. (2014). Zipf's word frequency law in natural language: A critical review and future directions. Psychonomic Bulletin \& Review, 21(5):1112-1130.
\bibitem{popescu2009word}  Popescu, I.  (2009). Word frequency studies, volume 64. Walter de Gruyter.
\bibitem{hall2009weka}   Hall, M., Frank, E., Holmes, G., Pfahringer, B., Reutemann, P., and Witten, I. H. (2009). The weka datamining software: an update. ACM SIGKDD explorations newsletter, 11(1):10-18.
\bibitem{zipf1935psycho}  Zipf, G. K. (1935). The psycho-biology of language. Houghton Mifflin Co, Boston.
\bibitem{zipf1949human}  Zipf, G. K.. (2016). Human behavior and the principle of least effort: An introduction to human ecology. Ravenio Books.
\bibitem{ioannides2003zipf} Ioannides, Y. M. and Overman, H. G. (2003). Zipf's law for cities: an empirical examination. Regional science and urban economics, 33(2):127-137.
\bibitem{gabaix2004evolution}  Gabaix, X. and Ioannides, Y. M. (2004). The evolution of city size distributions. Handbook of regional and urban economics , 4:2341-2378.
\bibitem{dimitrova2015primacy} Dimitrova, Z. and Ausloos, M. (2015). Primacy analysis of the system of Bulgarian cities. Open Physics,
13(1):218–225.
\bibitem{cerqueti2015evidence}  Cerqueti, R. and Ausloos, M. (2015). Evidence of economic regularities and disparities of Italian regions
from aggregated tax income size data. Physica A: Statistical Mechanics and its Applications , 421:187-207.
\bibitem{axtell2001zipf} Axtell, R. L. (2001). Zipf distribution of US firm  sizes. Science, 293(5536):1818-1820.
\bibitem{yoshi2004zipf} Fujiwara, Y. (2004). Zipf law in firms bankruptcy. Physica A: Statistical Mechanics and its Applications, 337(1-2), 219-230.
\bibitem{bottazzi2015zipf}  Bottazzi, G., Pirino, D., and Tamagni, F. (2015). Zipf law and the firm size distribution: a critical discussion of popular estimators. Journal of Evolutionary Economics, 25(3):585-610.
\bibitem{li2002zipf} Li, W. and Yang, Y. (2002). Zipf's law in importance of genes for cancer classification using microarray data. Journal of Theoretical Biology, 219(4):539-551
.\bibitem{levene2001zipf} Levene, M., Borges, J., and Loizou, G. (2001). Zipf's law for web surfers. Knowledge and Information Systems, 3(1):120-129.
\bibitem{maillart2008empirical}    Maillart, T., Sornette, D., Spaeth, S., and Von Krogh, G. (2008). Empirical tests of Zipf's law mechanism in open source Linux distribution. Physical Review Letters , 101(21):218701. 
\bibitem{manaris2005zipf} Manaris, B., Romero, J., Machado, P., Krehbiel, D., Hirzel, T., Pharr, W., and Davis, R. B. (2005). Zipf's law, music classification, and aesthetics. Computer Music Journal, 29(1):55-69.
\bibitem{zanette2006zipf} Zanette, D. H. (2006). Zipf's law and the creation of musical context. Musicae Scientiae, 10(1):3-18.
\bibitem{huang2008investigation}   Huang, S.-M., Yen, D. C., Yang, L.-W., and Hua, J.-S. (2008). An investigation of Zipf's law for fraud detection . Decision Support Systems, 46(1):70-83.
\bibitem{blasius2009zipf} Blasius, B. and Tonjes, R. (2009). Zipf's law in the popularity distribution of chess openings. Physical Review Letters, 103(21):218701.
\bibitem{pinto2012review} Pinto, C. M., Lopes, A. M., and Machado, J. T. (2012). A review of power laws in real life phenomena. Communications in Nonlinear Science and Numerical Simulation, 17(9):3558-3578.
\bibitem{montemurro2001beyond}  Montemurro, M. A. (2001). Beyond the Zipf-Mandelbrot law in quantitative linguistics. Physica A: Statistical Mechanics and its Applications , 300(3):567-578.
\bibitem{popescu2010zipf} Popescu, I.-I., Altmann, G., and Kohler, R. (2010). Zipf's law|another view. Quality \& Quantity, 44(4):713-731.
\bibitem{mandelbrot1953informational}Mandelbrot, B. (1953). An informational theory of the statistical structure of language. Communication theory, 84:486-502.
\bibitem{mandelbrot1961theory} Mandelbrot, B. (1961). On the theory of word frequencies and on related markovian models of discourse.
Structure of language and its mathematical aspects, 12:190-219.

\bibitem{fairthorne2005empirical}  Fairthorne, R. A. (2005). Empirical hyperbolic distributions (Bradford-Zipf-Mandelbrot) for bibliometric
description and prediction. Journal of Documentation , 61(2):171-193.
\bibitem{lavalette1996facteur} Lavalette, D. (1996). Facteur d'impact: impartialite ou impuissance. Report, INSERM U, 350:91405.
\bibitem{ausloos2016universal}  Ausloos, M. and Cerqueti, R. (2016). A universal rank-size law. PloS one , 11(11):e0166011.
\bibitem{i2006language} Ferrer- i Cancho, R. (2006). When language breaks into pieces a conflict between communication through isolated
signals and language. Biosystems, 84(3):242-253.
\bibitem{i2005consequences} Ferrer-i Cancho, R., Riordan, O., and Bollobas, B. (2005). The consequences of Zipf's law for syntax and symbolic reference. Proceedings of the Royal Society of London B: Biological Sciences, 272(1562):561-565.

\bibitem{jockers2014tex} Jockers, M. L. (2014). Text analysis with R for students of literature. Springer.
\bibitem{munzert2014automated} Munzert, S., Rubba, C., Meissner, P., and Nyhuis, D. (2014). Automated data collection with R: A practical
guide to web scraping and text mining. John Wiley \& Sons
\bibitem{mitkov2005oxford}
Mitkov R.
\newblock The Oxford handbook of computational linguistics.
\newblock Oxford University Press; 2005.
\bibitem{dkebowski2002zipf} Debowski, L. (2002). Zipf's law against the text size: a half-rational model. Glottometrics, 4:49-60.
 \bibitem{levenberg1944method}  Levenberg, K. (1944). A method for the solution of certain non-linear problems in least squares. Quarterly
of Applied Mathematics , 2(2):164-168.
\bibitem{marquardt1963algorithm} Marquardt, D. W. (1963). An algorithm for least-squares estimation of non-linear parameters. Journal of
the society for Industrial and Applied Mathematics, 11(2):431-441.
\bibitem{manning2008introduction}
Christopher~D M, Prabhakar R, Hinrich S.
\newblock Introduction to Information Retrieval.
\newblock Stanford University, Cambridge: Cambridge University Press; 2008.
\bibitem{lourakis2005brief} Lourakis, M. I. (2005). A brief description of the Levenberg-Marquardt algorithm implemented by levmar. Foundation of Research and Technology, 4(1) : 1-6.
\bibitem{bentz2014zipf}  Bentz, C., Kiela, D., Hill, F., and Buttery, P. (2014). Zipf's law and the grammar of languages: A quantitative
study of old and modern English parallel texts. Corpus Linguistics and Linguistic Theory, 10(2):175-211.
 \bibitem{shapiro1965analysis} Shapiro, S. S. and Wilk, M. B. (1965). An analysis of variance test for normality (complete samples).
Biometrika, 52(3/4):591-611.


\vskip0.5cm
 === XXXXX ===  speeches === XXXX ===
\vskip0.5cm
\bibitem{indianapolis}
{Miller Center}. Campaign speech in Indianapolis, Indiana.; 1932.
\newblock Available from:
  \url{https://millercenter.org/the-presidency/presidential-speeches/october-28-1932-campaign-speech-indianapolis-indiana}.

\bibitem{obamasou2016}
{Miller Center}. 2016 State of the Union Address; 2016.
\newblock Available from:
  \url{https://millercenter.org/the-presidency/presidential-speeches/january-12-2016-2016-state-union-address}.

\bibitem{chesterfam1981}
{Miller Center}. First Annual Message; 1981.
\newblock Available from:
  \url{https://millercenter.org/the-presidency/presidential-speeches/december-6-1981-first-annual-message}.

\bibitem{coolidgeia1925}
{Miller Center}. Inaugural Address; 1925.
\newblock Available from:
  \url{https://millercenter.org/the-presidency/presidential-speeches/march-4-1925-inaugural-address}.

\bibitem{obamarem2015}
{Miller Center}. Remarks in Eulogy for the Honorable Reverend Clementa Pickney;
  2015.
\newblock Available from:
  \url{https://millercenter.org/the-presidency/presidential-speeches/june-26-2015-remarks-eulogy-honorable-reverend-clementa}.

\bibitem{coolidgesam1928}
{Miller Center}. Sixth Annual Message; 1928.
\newblock Available from:
  \url{https://millercenter.org/the-presidency/presidential-speeches/december-4-1928-sixth-annual-message}.

\bibitem{reganfp1988}
{Miller Center}. Speech on Foreign Policy; 1988.
\newblock Available from:
  \url{https://millercenter.org/the-presidency/presidential-speeches/december-16-1988-speech-foreign-policy}.

\bibitem{clintonrem1996}
{Miller Center}. Remarks at the Democratic National Convention; 1996.
\newblock Available from:
  \url{https://millercenter.org/the-presidency/presidential-speeches/august-29-1996-remarks-democratic-national-convention}.

\bibitem{gwbushrem2004}
{Miller Center}. Remarks at the Democratic National Convention; 2004.
\newblock Available from:
  \url{https://millercenter.org/the-presidency/presidential-speeches/september-3-2004-remarks-republican-national-convention}.

\bibitem{lyndonpresconf1966}
{Miller Center}. Press Conference in the East Room; 1966.
\newblock Available from:
  \url{https://millercenter.org/the-presidency/presidential-speeches/july-20-1966-press-conference-east-room}.
  
  \bibitem{obamaaun2010}
{Miller Center}. Address to the United Nations; 2010.
\newblock Available from:
  \url{https://millercenter.org/the-presidency/presidential-speeches/september-23-2010-address-united-nations}.
  

\bibitem{regamrem1984}
{Miller Center}. Remarks Honoring the Vietnam War’s Unknown Soldier; 1984.
\newblock Available from:
  \url{https://millercenter.org/the-presidency/presidential-speeches/may-28-1984-remarks-honoring-vietnam-wars-unknown-soldier}.



\end{thebibliography}

\end{document}